\documentclass{article}

\usepackage{microtype}
\usepackage{graphicx}
\usepackage{subfigure}
\usepackage{booktabs} 

\usepackage{hyperref}

\usepackage[accepted]{icml2024}

\usepackage{amsmath}
\usepackage{amssymb}
\usepackage{mathtools}
\usepackage{amsthm}

\usepackage{xcolor,colortbl}
\usepackage{marvosym}
\usepackage{wrapfig}
\usepackage{caption}
\usepackage{subfigure}
\usepackage{multirow}
\usepackage{url}

\usepackage{array}
\usepackage{booktabs}
\usepackage[title]{appendix}
\usepackage{verbatim}
\usepackage{algorithm}
\usepackage{algorithmic}
\usepackage{setspace}
\renewcommand{\algorithmicrequire}{\textbf{Input:}} 
\renewcommand{\algorithmicensure}{\textbf{Output:}}
\usepackage{rotating}

\usepackage{bbding}
\usepackage{threeparttable}
\usepackage{pifont}
\usepackage{arydshln}

\usepackage{graphicx}
\usepackage{listings}
\usepackage{tcolorbox}
\usepackage{utfsym}
\tcbuselibrary{skins}
\usepackage{makecell}
\usepackage{changepage}

\usepackage[capitalize,noabbrev]{cleveref}

\theoremstyle{plain}

\theoremstyle{definition}

\theoremstyle{remark}

\usepackage[textsize=tiny]{todonotes}

\usepackage{multirow}
\usepackage{multicol}

\begin{document}

\twocolumn[
\icmltitle{BiLLM: Pushing the Limit of Post-Training Quantization for LLMs}



%
\begin{icmlauthorlist}
\icmlauthor{Wei Huang}{hku}
\icmlauthor{Yangdong Liu}{buaa}
\icmlauthor{Haotong Qin\textsuperscript{\Letter}}{eth}
\icmlauthor{Ying Li}{buaa}
\icmlauthor{Shiming Zhang}{hku}\\
\icmlauthor{Xianglong Liu}{buaa}
\icmlauthor{Michele Magno}{eth}
\icmlauthor{Xiaojuan Qi}{hku}
\end{icmlauthorlist}

\icmlaffiliation{buaa}{Beihang University}
\icmlaffiliation{eth}{ETH Zürich}
\icmlaffiliation{hku}{The University of Hong Kong}

\icmlcorrespondingauthor{Haotong Qin}{qinhaotong@buaa.edu.cn}

\icmlkeywords{Machine Learning, ICML}

\vskip 0.3in
]



\printAffiliationsAndNotice{}  

\begin{abstract}
Pretrained large language models (LLMs) exhibit exceptional general language processing capabilities but come with significant demands on memory and computational resources. As a powerful compression technology, binarization can extremely reduce model weights to a mere 1 bit, lowering the expensive computation and memory requirements. However, existing quantization techniques fall short of maintaining LLM performance under ultra-low bit-widths. In response to this challenge, we present BiLLM, a groundbreaking 1-bit post-training quantization scheme tailored for pretrained LLMs. Based on the weight distribution of LLMs, BiLLM first identifies and structurally selects salient weights, and minimizes the compression loss through an effective \textit{binary residual approximation} strategy. Moreover, considering the bell-shaped distribution of the non-salient weights, we propose an \textit{optimal splitting search} to group and binarize them accurately. BiLLM, for the first time, achieves high-accuracy inference (e.g. 8.41 perplexity on LLaMA2-70B) with only \textbf{1.08-bit} weights across various LLM families and evaluation metrics, outperforms SOTA quantization methods of LLM by significant margins. Moreover, BiLLM enables the binarization process of a 7-billion LLM within 0.5 hours on a single GPU, demonstrating satisfactory time efficiency.
Our code is available at \href{https://github.com/Aaronhuang-778/BiLLM}{https://github.com/Aaronhuang-778/BiLLM}.
\end{abstract}

\section{Introduction}

Recently, large language models (LLMs) based on transformers~\cite{vaswani2017attention} have garnered significant attention in natural language processing. Pre-trained LLMs such as OPT~\cite{zhang2022opt} and LLaMA~\cite{touvron2023Llama}, have demonstrated excellent performance across a range of evaluation benchmarks. However, LLMs pose substantial challenges in deployment on memory-constrained devices due to their immense parameter size and computation requirements. For instance, the widely-used LLaMA2-70B~\cite{touvron2023Llama2} model, with its 70 billion parameters, requires 150 GB of storage in half-precision (FP16) format. This necessitates at least two A100 GPUs, each with 80 GB of storage space, for inference.

\begin{figure}[!t]
\centerline{\includegraphics[width=1.\columnwidth]{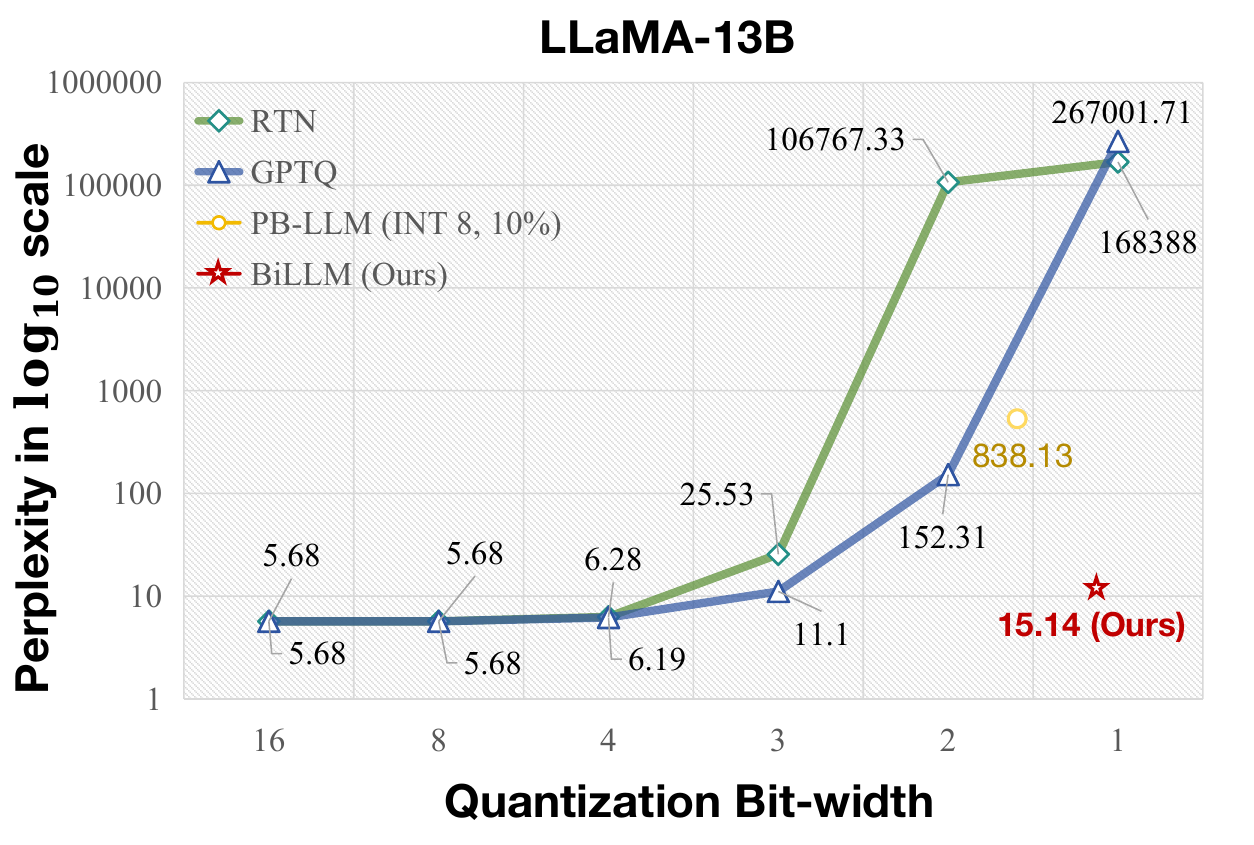}}
\caption{Perplexity of LLaMA-13B on WikiText2 under different bit-widths. Round-to-nearest (RTN), GPTQ, and PB-LLM (10\% weight of INT8) suffer accuracy loss at ultra-low bits, facing the sharply increasing perplexity ($\downarrow$). \textit{BiLLM} demonstrates exceptional performance under binarization.}
\label{binary_results}
\end{figure}

Model quantization has emerged as a highly effective technology for compressing neural networks, thereby reducing the model size of LLMs and substantially saving GPU memory consumption~\cite{dettmers2022llm}. Current quantization techniques primarily fall into Quantization-Aware Training (QAT) and Post-Training Quantization (PTQ). QAT involves fine-tuning and retraining during the quantization process, while PTQ significantly streamlines the computation by eliminating back-propagation, enabling a faster quantization process and promoting the practicality of quantization~\cite{frantar2022gptq,shang2023pb,lin2023awq}. Given the deep structures and numerous parameters of LLMs, PTQ stands out for its ability to rapidly perform the quantization process, especially on time and resource-constrained scenarios~\cite {zhu2023survey}.

Despite the success of previous PTQ methods in 8-bit and 4-bit quantization~\cite{dettmers2022llm,dettmers2023spqr,frantar2022gptq,xiao2023smoothquant,frantar2022optimal}, the expanding size of LLMs demands more aggressive quantization approaches~\cite{shang2023pb}. Neural network binarization, which reduces the weight bit-width to only 1 bit, is a promising approach~\cite{helwegen2019latent,qin2020forward,qin2023bibench}. However, as depicted in Figure~\ref{binary_results}, current advanced PTQ methods for LLMs exhibit a performance collapse under ultra-low bit ($\leqslant$3 bits) quantization. This phenomenon can be attributed to the significant difference between quantized and original weights. Even the recent binary PTQ method for LLMs, PB-LLM~\cite{shang2023pb}, only maintains a perplexity metric of around 800 with an average weight of 1.7 bits. This observation underscores the challenges existing PTQ methods face in promoting the weight binarization of LLMs.

\begin{figure}[!t]
\includegraphics[width=1\columnwidth]{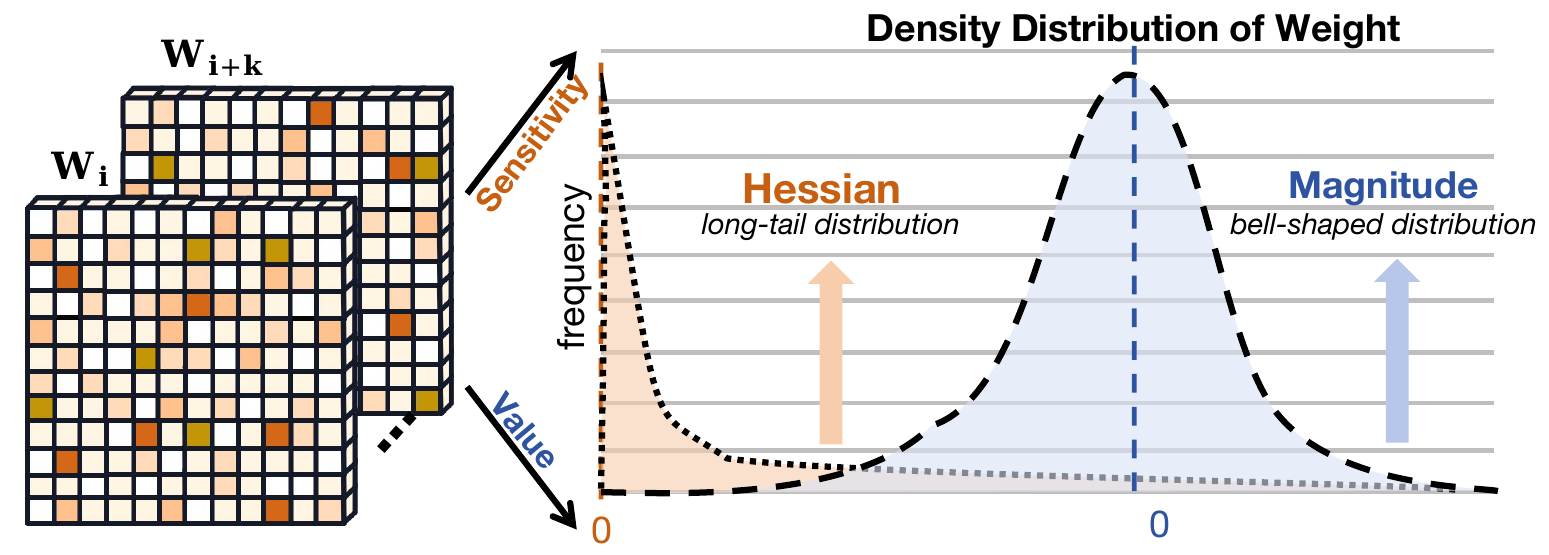}
\caption{The Hessian metrics (sensitivity) and magnitude (value) of weights in LLMs.
The weights of different layers in LLMs are characterized by bell-shaped distribution, accompanied by a few salient values.}
\label{distribution}
\end{figure}

In pursuit of this goal, we conducted an empirical study to analyze the distribution of pre-trained weights in LLMs. The findings derived from this study are presented in Appendix \ref{more_dis}, revealing two key observations:
\begin{itemize}
\item The second-order Hessian matrix of weights demonstrates an \textbf{exceptionally long-tail distribution} and is often used to measure the importance of weight elements in neural networks~\cite{lecun1989optimal,dong2019hawq}. As depicted in Figure~\ref{distribution}, a small fraction of weights elements possesses significantly high Hessian values, substantially influencing the layer output. In contrast, most Hessian values cluster around 0.

\item The density distribution of weight magnitudes in LLMs follows a \textbf{bell-shaped pattern}. This bell-shaped distribution exhibits a significant resemblance to both the Gaussian or Laplace distribution in terms of its characteristics~\cite{blundell2015weight}. Figure~\ref{distribution} illustrates that most weight values cluster around zero with a non-uniform bell-shaped distribution.
\end{itemize}
The above implies: 
a) A minority of weights play an important role in LLMs, whereas the majority of weights exhibit characteristics of redundancy~\cite{shang2023pb,dettmers2023spqr}; 
b) With the most aggressive bit-width, binarization incurs most severe error among quantization under bell-shaped distributions in LLMs~\cite{jacob2018quantization}.

\begin{figure*}[!ht]
\includegraphics[width=1\textwidth]{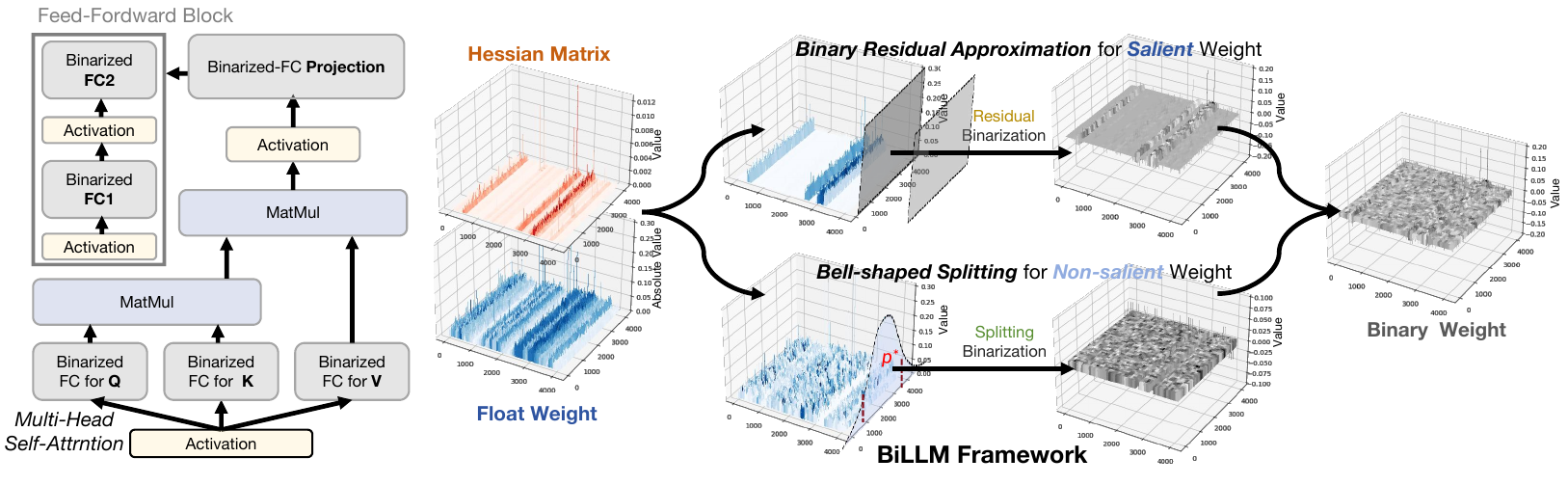}
\caption{Schematic of the PTQ binarization framework for LLMs. The left side shows the structure of the Transformer block after binarization. The right side shows the binarization process of \textit{BiLLM}, which consists of two parts, \textit{Residual Approximation} for salient weights and \textit{Bell-shaped Splitting} for non-salient weights.}
\label{main_billm}
\end{figure*}

Motivated by the above observation, we propose a novel 1-bit PTQ framework for LLMs, namely \textit{BiLLM}, incorporating two core designs to achieve highly accurate weight binarization. 
First, guided by the Hessian-based metric, we select the salient weights structurally (Figure~\ref{main_billm} upper-right) to achieve a trade-off between accuracy and storage savings and develop a residual approximation to maximize the restoration of salient weights with highly dynamic range. 
Second, for the remaining non-salient weights (Figure~\ref{main_billm} lower-right), we design an optimal splitting binarization strategy, where a meticulous search process is applied to determine an optimal break-point for weight distribution and binarization of the segments is then processed separately to minimize binarization errors. 
Moreover, \textit{BiLLM} incorporates error compensation on a block-wise basis by default following existing common practices~\cite{frantar2022gptq,shang2023pb}, which further reduces quantization error.

Extensive experiments demonstrate that \textit{BiLLM} achieve the state-of-the-art (SOTA) performance for LLMs across multiple LLM families on various evaluation metrics, and first achieves extremely compact 1.07$\sim$1.11 bit-width in average for the PTQ binarization. For example, on the Wikitext2\cite{merity2016pointer} metric, \textit{BiLLM} achieved perplexities of 8.49 and 8.41 with only 1.08-bit weights on LLaMA-65B~\cite{touvron2023Llama}and LLaMA2-70B~\cite{touvron2023Llama2}, respectively, even surpassing the 9.34 performance of the FP16 OPT-66B~\cite{zhang2022opt}.

\section{Related Works}\label{section2}

\subsection{Large Language Model Quantization}
Quantization maps high-precision parameters to a discrete range. This method, which compresses parameters without altering the model structure, effectively reduces the storage and computational overhead of deep neural networks. Recent work has successfully applied QAT and PTQ to LLMs. QAT, through a quantization-aware retraining strategy, better preserves the performance of quantized models. LLM-QAT~\cite{liu2023llm} addressed data barrier issues in QAT training through data-free distillation. However, for LLMs with extremely large parameter sizes, the cost of retraining is prohibitively high and inefficient. Therefore, techniques such as QLoRA~\cite{dettmers2023qlora} focus on parameter-efficient fine-tuning (PEFT) methods for quantizing LLMs, enhancing the efficiency of QAT. Nevertheless, even these efficient fine-tuning quantization strategies require over 24 hours of GPU time. 

Therefore, the PTQ strategy has become a significant option for quantizing LLMs efficiently. Works like BRECQ~\cite{li2021brecq}, ZerqQuant~\cite{yao2206efficient} and LLM.int8()~\cite{dettmers2022llm} enhance quantization accuracy by adding additional grouping labels for custom quantization blocks. Other studies adopt a feature segmentation strategy, such as PB-LLM~\cite{shang2023pb} and SpQR~\cite{dettmers2023spqr}. They preserve the bit-width of outlier features or those with higher quantization errors to FP16 or INT8, mitigating the precision loss due to quantization. GPTQ~\cite{frantar2022gptq} employs a more precise quantization framework, reducing the block quantization errors of LLMs through Hessian-based second-order error compensation~\cite{frantar2022optimal}, achieving commendable performance in low-bits (4 bits) quantization. Smoothquant~\cite{xiao2023smoothquant} introduces a strategy of scaling weight and activation outliers to simplify quantization. Subsequently, AWQ~\cite{lin2023awq} and OWQ~\cite{lee2023owq} also proposed scale transformations of more crucial weight channels for activation features, preserving their information representation capacity. 

\subsection{Network Binarization}

Binarized compression can quantize parameters to only 1 bit, expressed as $\pm$1. In forward propagation, the sign function is used to binarize the original parameter tensor:
\begin{equation}\label{eq1}
    \mathbf{W}_b = \alpha \cdot \operatorname{sign}(\mathbf{W}_{f}),
\end{equation}
\begin{equation} \label{sign}
    \operatorname{sign}(x) = 
    \begin{cases}
    1& \text{if $x \geq 0$},\\
    -1& \text{others}.
    \end{cases}
  \end{equation}
where $\mathbf{W}_f \in \mathbb{R}^{n\times m}$ is the full precision weight and $\mathbf{W}_b\in \mathbb{R}^{n\times m}$ is the binarized output. $n$ and $m$ represent the size of the weight matrix. $\alpha$ denotes the scaling factor~\cite{courbariaux2016binarized}. Binarization usually uses the channel-wise scale~\cite{rastegari2016xnor,qin2023bibench}, so $\alpha \in \mathbb{R}^n$. 

Most previous binarization works adopt a framework based on QAT for quantization~\cite{qin2023bibench}. Straight through estimator (STE)~\cite{bengio2013estimating} is deployed to address the issue of gradient vanishing caused by the $\operatorname{sign}(\cdot)$ function. Binary Weight Network (BWN)~\cite{rastegari2016xnor} was initially proposed for executing neural network computations by binarizing weights and using full-precision activations, while XNOR-Net~\cite{rastegari2016xnor} extends this approach by binarizing both weights and activations. Both methods minimize quantization errors through dynamic searching of $\alpha$. DoReFa-Net~\cite{zhou2016dorefa} further expands upon XNOR-Net, employing quantized gradients to accelerate network training.  Group segmentation is also applied in binarization tasks, with Syq~\cite{faraone2018syq} utilizing network weight to the small size of groups for minimizing binarization errors.

Based on the successful application of binarization in Transformers~\cite{wang2023bitnet} and Bert~\cite{qin2022bibert}, we believe that the binarization of LLMs is filled with potential. PB-LLM~\cite{shang2023pb} investigates the impact of binarized QAT and PTQ strategies on LLMs, but it is necessary to retain a significant proportion (over 30\%) of the weights at 8 bits to enable LLMs to produce reasonable answers. Due to the presence of a large amount of INT8, LLMs still have a relatively high average bit-width. To address this issue, we proposed \textit{BiLLM}, which aims to push the limit of PTQ binarization for LLMs.

\section{Method}
To achieve accurate binarization of LLMs, our approach is designing distinct binarization strategies for salient and non-salient weights. We first introduce the selection rules for salient weights and their binarization strategies in Section \ref{sec_salient}. Then, we elaborate on the distribution-based binarization strategy for non-salient weights in Section \ref{sec_unsalient}.
\subsection{Salient Weight Binarization for LLMs} \label{sec_salient}

In deep neural networks, not all parameters carry equal significance. Utilizing solely the magnitude of the weights can not fully capture the impact of each element on the model's performance. The Hessian metric serves as a common benchmark for detecting parameter sensitivity~\cite{dong2019hawq,dettmers2023spqr,dettmers2022llm}.
We thus leverage the Hessian matrix to assess the salience of parameters in each under-binarized layer. We implement an optimized computation process to derive weight sensitivity, which allows us to obtain the importance metric of parameters without compromising efficiency:
\begin{equation}
s_i = \frac{w_i^2}{[\mathbf{H}^{-1}]_{ii}^2},
\end{equation}
where $\mathbf{H}$ represents the Hessian matrix of each layer, and $w_i$ represents the original value of each element. In the following section, $s_i$ serves as a criterion for assessing the significance of weight elements and is used as a feature indicator for structured selection.

\textbf{Structural Searching Selection.} Utilizing an unstructured selection enables the coverage of all salient elements. However, it requires the implementation of an additional 1-bit bitmap index~\cite{chan1998bitmap}, leading to increased average bit-width. This balance is inefficient, especially for Hessian outlier weights that constitute a mere 1-5\% of the total~\cite{yao2023comprehensive}. In our analysis of sensitivity distribution within LLMs, we discovered that the majority of the weights' sensitive Hessian values are predominantly concentrated in specific columns or rows (Appendix \ref{more_dis}). This pattern is attributed to the convergence effects inherent in the multi-head self-attention mechanism of these models and further motivates us to implement a structured approach for selecting salient weights, for reducing the additional bitmap. Given that \textit{BiLLM} employs a per-channel (or per-row) type of binarization, we determine salience through a per-column segmentation on the whole weight matrix.

We organize the column salience in descending order and introduce an optimized search algorithm aimed at minimizing quantization error, which in turn determines the number of columns within the salient group. To elaborate on this methodology, we initially define the objective of binarization quantization, grounded on Equation~\eqref{eq1}:
\begin{equation}\label{eq4}
\mathop{\arg\min}\limits_{\alpha, \mathbf{B}}||\mathbf{W} - \alpha \mathbf{B}||^2,
\end{equation}
where $\mathbf{B}\in\{-1, +1\}^{n\times k}$, $k$ is the number of selected columns. The problem~\cite{rastegari2016xnor} of optimal $\alpha$ and $\mathbf{B}$ can simply be solved as $\alpha = \frac{||\mathbf{W}||_{\ell1}}{n\times k}$ and $\mathbf{B}=\operatorname{sign}(\mathbf{W})$. Then, the optimization function for selecting salient columns is defined as:
\begin{equation}\label{eq_column_search}
\mathop{\arg\min}\limits_{\mathbf{W}_{\text{uns}}}||\mathbf{W} - (\alpha_{\text{sal}} \operatorname{sign}(\mathbf{W}_{\text{sal}}) \cup \alpha_{\text{uns}} \operatorname{sign}(\mathbf{W}_{\text{uns}}))||^2,
\end{equation}
where $\mathbf{W}_{\text{sal}}$ denotes the column-wise combination of original weight and $\mathbf{W}_{\text{uns}}$ is the left non-salient part. We can easily get that $\mathbf{W} = \mathbf{W}_{\text{sal}} \cup \mathbf{W}_{\text{uns}}$, so the only variable parameter is the number of rows in $\mathbf{W}_{\text{sal}}$. 


\begin{figure}[!t]
\centerline{\includegraphics[width=1\columnwidth]{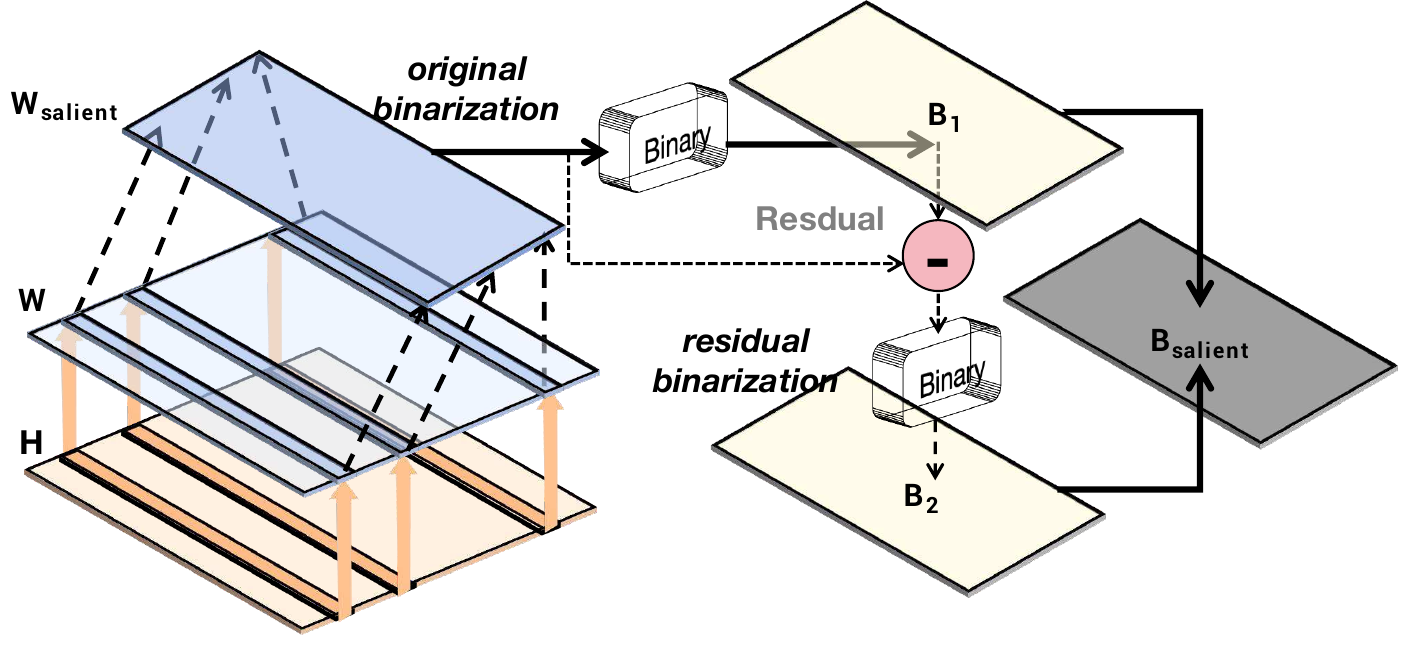}}
\vspace{-0.1in}
\caption{Illustration of salient weight binarization. The $\mathbf{B}_1$ binarized from salient weight is made into a residual with the original value and then binarized again to obtain $\mathbf{B}_2$.}
\label{structured_residual}
\end{figure}

\textbf{Binary Residual Approximation.}
Salient weights are limited in quantity, yet exhibit significant variance when aggregated. Direct preservation of these weights in INT8 or FP16 formats leads to an increase in the average weight bits, undermining the compressive benefits of binarization. Traditional binarization methods for salient weights, however, result in substantial quantization errors. 
To that end, we develop a residual approximation approach for binarizing salient weights. Contrary to the comprehensive high-order quantization~\cite{li2017performance} applied to the entire weight matrix, our technique minimizes binarization error through a second-order approximation of merely a select subset of salient weights. This method guarantees the precision of salient weights while simultaneously decreasing bit-width overhead. As illustrated in Figure~\ref{structured_residual}, this approach incorporates a recursive computation strategy for weight binarization compensation, applying a subsequent binarization process to the residuals remaining after the initial binary process. Building upon Equation~\eqref{eq4}, we propose a redesigned residual approximation optimization specifically for salient weights, which is defined as follows:
\begin{equation}\label{eq5}
    \left\{
         \begin{array}{lr}
          \alpha_o^*, \mathbf{B}_o^* = \mathop{\arg\min}\limits_{\alpha_o, \mathbf{B}_o}||\mathbf{W} - \alpha_o \mathbf{B}_o||^2), \\
          \alpha_r^*, \mathbf{B}_r^* = \mathop{\arg\min}\limits_{\alpha_r, \mathbf{B}_r}||(\mathbf{W} - \alpha_o^* \mathbf{B}_o^*) - \alpha_r \mathbf{B}_r||^2), \\
         \end{array}
    \right.
\end{equation}
where $\mathbf{B}_o$ represents the original binary tensor, while $\mathbf{B}_r$ denotes the residual binarized matrix with the same size as $\mathbf{B}_o$. We efficiently solve for the two binarized optimization objectives using the same solution method as in Equation~\eqref{eq4}. Ultimately, we arrive at the following approximation:
\begin{equation}\label{eq6}
    \mathbf{W} \approx \alpha_o^* \mathbf{B}_o^* + \alpha_r^* \mathbf{B}_r^*.
\end{equation}
It can be easily proven that the residual approach of Equation~\eqref{eq6} has a lower quantization error than the direct one of Equation~\eqref{eq4}. We define the residual binarization error $\mathcal{E}$:
\begin{equation}\label{eq7}
    \mathcal{E}_{rb} = ||\mathbf{W} - \alpha_o^* \mathbf{B}_o^* - \alpha_r^* \mathbf{B}_r^*||^2.
\end{equation}
The original binarized quantization error is calculatde as $||\mathbf{W} - \alpha_o^* \mathbf{B}_o^*||^2$ by Equation~\eqref{eq4}, and from the second sub-equation of Equation~\eqref{eq5} we can get that loss $\mathcal{E}_{rb} \leq ||\mathbf{W} - \alpha_o^* \mathbf{B}_o^*||^2$. Therefore, through the method of residual approximation, we are able to further reduce the binary quantization error of salient weights with ultra-low bit-width storage compared to retaining salient weights at 8 or 16 bits.

\begin{figure}[!t]
\centerline{\includegraphics[width=1\columnwidth]{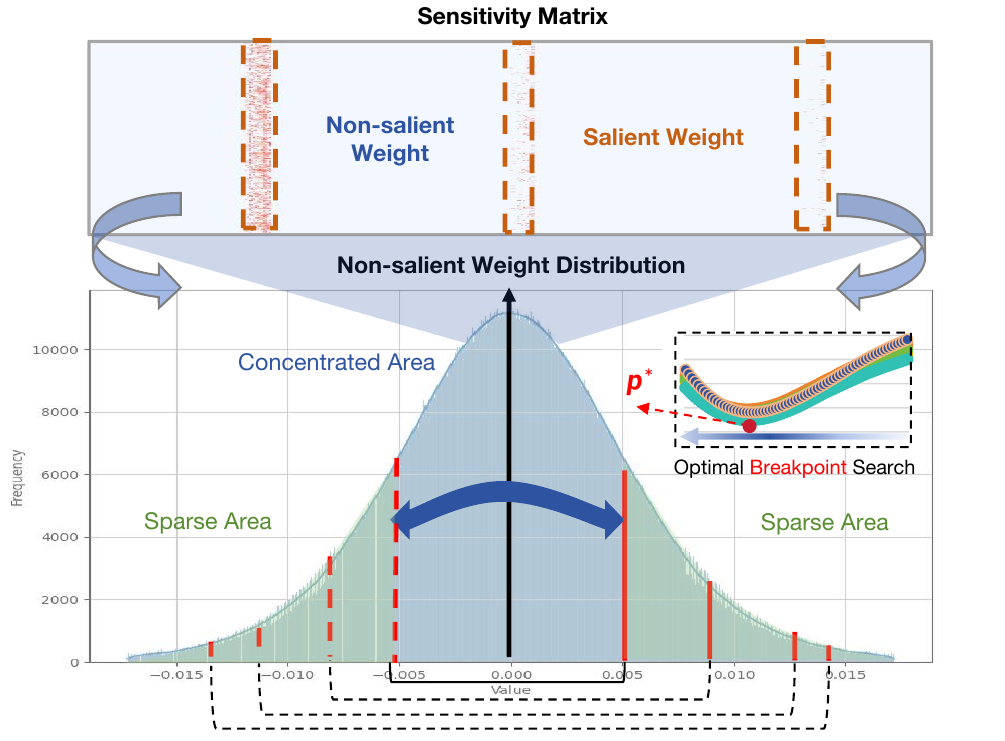}}
\vspace{-0.1in}
\caption{Distribution and splitting schematic of the $4^{th}$ projection layer in LLaMA2-7B. The top 5\% of the Hessian elements are orange, and the optimal break-point divides the non-salient weights into sparse and concentrated areas.}
\label{gaussian_curve}
\end{figure}

\subsection{Bell-shaped Distribution Splitting for Binarization}\label{sec_unsalient}
Following the removal of salient weights, the remaining weights maintain a bell-shaped distribution, which becomes closer to symmetric with the exclusion of salient weights' impact, as depicted in Figure~\ref{gaussian_curve}. Binary quantization, representing an extreme form of uniform quantization, encounters more loss in the presence of non-uniform distributions. A practical approach involves the group-wise quantization~\cite{park2018value,fang2020post,jain2019biscaled} of weights according to their distribution. Balancing between quantization accuracy and compression efficiency, we identify a single break-point within the distribution. As shown in Figure~\ref{gaussian_curve}, this partition divides the non-salient bell-shaped distribution into two categories: the sparse area and the concentrated area.

The segmentation process identifies a break-point that categorizes non-salient weights into two groups: $A_c[-p, p]$  for concentrated weights and $A_s[-m, -p]\cup[p, m]$ for sparse weights, where signifies the maximum extent of non-salient weights. We then apply binarization to both $A_c$ (concentrated) and $A_s$ (sparse). To determine the optimal break-point $p^*$, we assume that the non-salient weights possess a symmetrical probability density function (PDF)-$g(x)$ over the bounded domain $[-m, m]$, with the properties $g(x) = g(-x)$. Then the mean squared quantization error of binarization is defined as:
\begin{equation}  
\theta_{q}^2 = \int_{-m}^{0} (-\alpha - x)^2 g(x) dx + \int_{0}^{m} (\alpha - x)^2 g(x) dx.
\end{equation}
Since $g(x)$ is a symmetric function, the above formula is simplified to:
\begin{equation}\label{msqe_b}
    \theta_{q}^2 = 2\int_{0}^{m} (\alpha - x)^2 g(x) dx.
\end{equation}
Then, the break-point $p$ divides the non-salient weights into two parts. According to the Equation~\eqref{msqe_b}, under the discontinuous weight distribution, we get a new binary quantization error:
\begin{equation}\label{msqe_b1}
 \theta_{q,p}^2 = || \mathbf{W}_s - \alpha_s\mathbf{B}_s||^2 + || \mathbf{W}_c - \alpha_c\mathbf{B}_c||^2,
\end{equation}
where $\mathbf{W}_s$ and $\mathbf{W}_c$ denote the weights of the sparse and concentrated area, respectively. $\mathbf{B}_s$ and $\mathbf{B}_c$ were calculated from Equation~\eqref{sign}, $\alpha_s$ and $\alpha_c$ are the binarization scales, determined by Equation~\eqref{eq4}:
\begin{equation}\label{eq_scale_area}
    \alpha_s = \frac{1}{n_s}||\mathbf{W}_s||_{\ell1},  
    \alpha_c = \frac{1}{n_c}||\mathbf{W}_c||_{\ell1},  
\end{equation}
where $n$ represents the number of weight elements in each area. Therefore, the problem function is only related to $p$, and our target to find the optimal $p^*$ can be defined as:
\begin{equation}  \label{optimal}
p^* = \mathop{\arg\min}_{p} (\theta_{q,p}^2).
\end{equation}
When the remaining weights follow an ideal Gaussian distribution, Equation~\eqref{msqe_b1} is demonstrated to be a convex function with a global minimum, as evidenced in prior studies~\cite{fang2020post,you2010audio}. Nonetheless, the actual distribution of non-salient weights, while bell-shaped, diverges from the ideal Gaussian model. Simultaneously, we retain the block-wise compensation strategies of GPTQ~\cite{frantar2022gptq} and OBC~\cite{frantar2022optimal} to offset quantization errors, which could change the distribution of weights. In response, we employ a percentile search method to identify the optimal break-point based on the objective function outlined in Equation~\eqref{optimal}. This percentile search strategy is efficient and straightforward, completing the binarization process for a 7B LLM within merely 30 minutes. Furthermore, our findings indicate that despite the deviation of non-salient weights from the ideal Gaussian distribution, the error curve associated with the search process still exhibits convex properties (as detailed in Appendix \ref{ap_seg}), confirming the feasibility of pinpointing the optimal break-point.

\subsection{Pipeline of \textit{BiLLM}}
As depicted in Figure~\ref{main_billm} left, \textit{BiLLM} primarily performs binary quantization on all Linear weights within the Transformer blocks. This section introduces the detailed pipeline of \textit{BiLLM}. 

\textbf{Binarization Workflow.}
We first deploy the structural search of salient columns and a residual approximation binarization for salient columns. The process of salient columns incurs additional weight bits due to the search proportion and residual mechanism. Table \ref{salient_searching_bits} presents the extra bits generated in some LLMs~\cite{zhang2022opt,touvron2023Llama, touvron2023Llama2}. It can be observed that the searching and residuals bring only about \textbf{0.1} additional weight bits. Then, for these non-uniformly distributed weights, we use a split binarization strategy searching optimal $p^*$. The concentrated area and the sparse area are binarized separately. This part incurs the cost of an additional 1 bit for hardware group identification, but the computing parameters are still compressed to 1 bit. By retaining only block-wise compensation\cite{frantar2022gptq,frantar2022optimal} and eliminating column-wise quantization error compensation, we further enhance the efficiency of PTQ and ensure the effectiveness of distribution exploration. Algorithm \ref{alg1} illustrates the complete process of \textit{BiLLM}, and detailed implementation of \textit{BiLLM} is shown in Appendix \ref{func}.

\begin{figure}[!t]
\centerline{\includegraphics[width=1.\columnwidth]{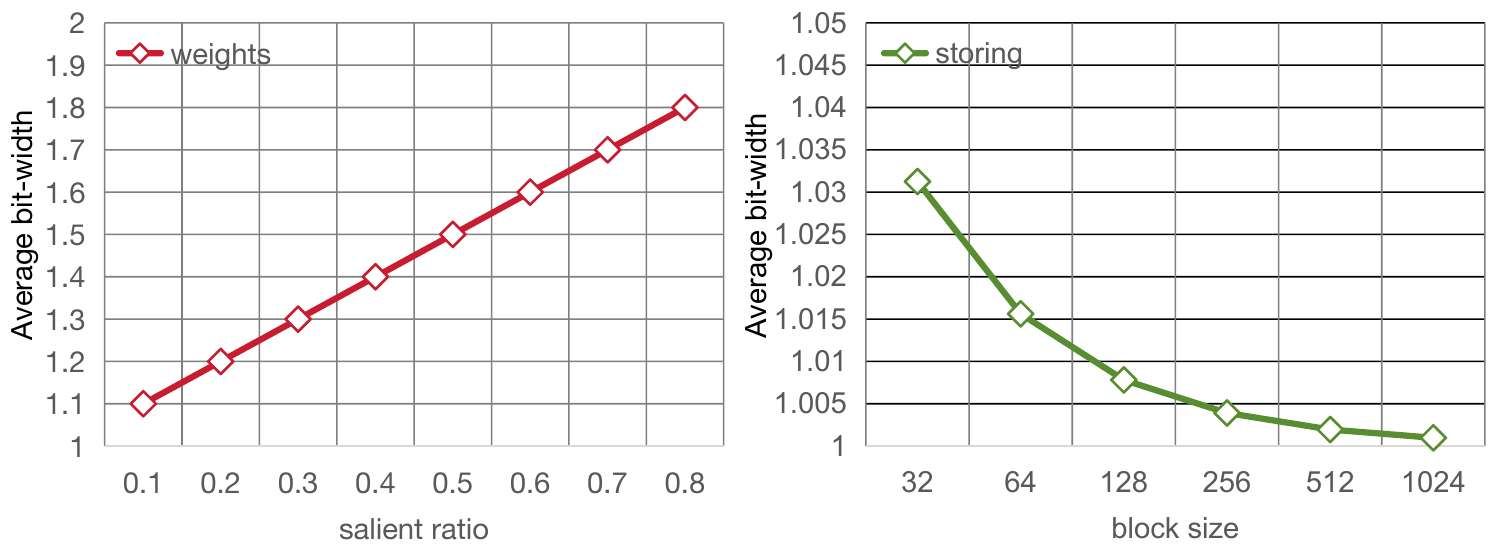}}
\vspace{-0.1in}
\caption{Weights and hardware overhead changes on Llama-7B. The left shows the calculation parameters as a function of the significant weight ratio; the right shows the hardware overhead as a function of the block.}
\label{storing}
\end{figure}

\begin{table}[!t]
\centering
\setlength{\tabcolsep}{3.mm}
\caption{
Average bit results from structural searching and residual binarization of OPT, LLaMA, and LLaMA2.
}
\vspace{-0.1in}
\begin{tabular}{lrrrc}
\toprule
\textbf{Model} & \textbf{7B} &  \textbf{13B} & \textbf{30B} & \textbf{66B/65B/70B}*  \\

\midrule
OPT & 1.10	&1.12	&1.12	&1.13 \\
LLaMA & 1.09	&1.09	&1.10	& 1.10 \\
LLaMA2 & 1.07&	1.08& N/A &	1.09 \\

\bottomrule

\end{tabular}
\begin{tablenotes}
        \footnotesize
        \item *: OPT-66B, LLaMA-65B and LLaMA2-70B.
\end{tablenotes}
\label{salient_searching_bits}
\end{table}
\textbf{Extra Storing Bits.}
The extra bits is acceptable under the binary weight quantization of \textit{BiLLM}. The weight parameters and additional hardware overhead are as follows:
\begin{equation}
    \left\{
         \begin{array}{lr}
         N_{\text{param}} = 2 \times r_{\text{salient}} + 1 \times (1-r_{\text{salient}}), \\
         N_{\text{storing}} =  1 + \dfrac{1}{b_{\text{size}}}, \\
         \end{array}
    \right.
\end{equation}
where $r_{salient}$ signifies the proportion of salient weights and $b_{size}$ denotes the block size in OBC compensation, with 1 bit allocated for marking the division of non-salient weights. $\frac{1}{b_{size}}$ represents the identifier for the structured column of salient weights. For example, a 10\% structural selection along with an OBC compensation of size 128 was employed. This results in a weight parameter bit-width of \textbf{1.1} bits and a hardware flag bit-width of 1.008 bits. Figure~\ref{storing} illustrates the weight overhead for different proportions and block sizes. It is important to note that flag weights do not participate in the computation; actual calculations are executed solely with parameter weights. Therefore, additional hardware identification bits do not affect the acceleration effect of binary quantization.
\begin{algorithm}[!h]
\caption{Main Framework of BiLLM: Inner details of each function are shown in Algorithm \ref{alg2} }
\label{alg1}
func $\operatorname{BinaryLLM}$($\mathbf{W}$, $\mathbf{X}$, $\beta$, $\lambda$)\\
{\bf Input:} $\mathbf{W} \in \mathbb{R}^{n\times m}$ - weight matrix \\
\hspace*{0.43in}$ \mathbf{X} \in \mathbb{R}^{r\times d}$ - calibration data \\
\hspace*{0.43in}$\beta$ - block size\\
\hspace*{0.43in}$\lambda$ - hessian regularizer\\
{\bf Output:} $ \mathbf{B}$ - binarized  weights
\begin{algorithmic}[1]
\STATE $\mathbf{H} \coloneqq 2\mathbf{X}\mathbf{X}^\top \ $ //  $\ell^2$ error hessian matrix 
\STATE $\mathbf{H}^c \coloneqq $ Cholesky$({(\mathbf{H} + \lambda \mathbf{I})}^{-1})$
\STATE $\mathbf{B} \coloneqq 0_{n\times m}$
\FOR{$b=0, \beta, 2\beta,...,N$} 
    \STATE $\mathbf{W}^b \coloneqq \mathbf{W}_{:, b:b+\beta}$
    \STATE $row_{s}\{\cdot\}\coloneqq \operatorname{salient} (\mathbf{W}_{:, b:b+\beta}, \mathbf{H}^c)$
    \STATE $\Tilde{\mathbf{B}}_1 \coloneqq \operatorname{res\_approximation} (\mathbf{W}_{:, j \in \{ row_{s}\}}^b)$
    \STATE $p^* \coloneqq \operatorname{seg\_search} (\mathbf{W}_{i,j \notin \{row_{s}\}}^b)$
    \STATE $\Tilde{\mathbf{B}}_2 \coloneqq \operatorname{binary}(\mathbf{W}_{|w_{i,j}| \leq p^*, j \notin \{row_{s} \}}^b)$
    \STATE $\Tilde{\mathbf{B}}_3 \coloneqq \operatorname{binary}(\mathbf{W}_{|w_{i,j}| > p^*, j \notin \{ row_{s} \}}^b)$
    \STATE $\mathbf{B}_{:, b:b+\beta} \coloneqq \Tilde{\mathbf{B}}_1+ \Tilde{\mathbf{B}}_2 + \Tilde{\mathbf{B}}_3$
    \STATE $\mathbf{E} \coloneqq (\mathbf{W}_{:, b:b+\beta} - \mathbf{B}_{:, b:b+\beta}) / \mathbf{H}^c_{bb:b+\beta b+\beta}$
    \STATE $\mathbf{W}_{:, b+\beta:} \coloneqq \mathbf{W}_{:, b+\beta:} - \mathbf{E} \cdot \mathbf{H}^c_{b:b+\beta, b+\beta:}\ $ // block-wise OBC 
\ENDFOR
\STATE {\bf return}  $\mathbf{B}$

\end{algorithmic}
\end{algorithm}

\begin{table*}[!t]
\centering
\setlength{\tabcolsep}{2.6mm}
\caption{
Perplexity of RTN, GPTQ, PB-LLM, and BiLLM on OPT Family. The columns represent the perplexity results on Wikitext2 datasets with different model sizes.
}
\vspace{-0.1in}
\begin{tabular}{lcrrrrrrr}
\toprule
\textbf{Method} & \makecell{\textbf{Block} \\ \textbf{Size}}  & \makecell{\textbf{Weight} \\ \textbf{Bits}}  & \textbf{1.3B} & \textbf{2.7B} & \textbf{6.7B} & \textbf{13B} & \textbf{30B} & \textbf{66B} \\

\midrule

Full Precision & - & 16.00 & 14.62 & 12.47 & 10.86 & 10.13 & 9.56 & 9.34\\
\cline{1-9}
RTN & - & 3.00  & 13337.38 & 15594.72 & 5797.32 & 3357.01 & 1566.00 & 6126.09\\
GPTQ & 128 & 3.00  & 20.97 & 16.88 & 14.86 & 11.61 & 10.27 & 10.51\\
\hdashline
RTN & - & 2.00 & 11272.65 & 9505.76 & 28363.14 & 194086.78 & 169616.47 & 1165864.25\\
GPTQ  & 128& 2.00  & 115.17 & 61.59 & 50.19 & 21.36 & 15.71 & 82.10\\ 
RTN & - & 1.00  & 17165.72 & 36516.69 & 11550.91 & 6986.35 & 6485.99 & 184796.30\\
GPTQ  & 128& 1.00 & 14884.73 & 14144.58 & 10622.81 & 15196.96 & 12478.37 & 13106.45\\
PB-LLM $\dagger$ & 128 & 1.70 & 265.52 & 124.35 & 105.16 & 81.92 & 25.14 & 29.09\\
\textbf{BiLLM} $\ddagger$& 128 & \textbf{1.11}  & \textbf{69.97} & \textbf{49.55} & \textbf{35.36} & \textbf{18.82} & \textbf{12.71} & \textbf{12.06}\\
\bottomrule

\end{tabular}
\begin{tablenotes}
        \footnotesize
        \item  -: Vanilla RTN conducts layer-wise quantization. $\dagger$: PB-LLM selects 10\% elements in the original tensor as salient weights based on Hessian. $\ddagger$: BiLLM uses structural searching for salient weights. The table gives the average bit-width of the OPT family.
\end{tablenotes}
 \label{exp_opt}
\end{table*}

\section{Experiments}
\subsection{Setup}
We deploy \textit{BiLLM} within the Pytorch~\cite{paszke2019pytorch}-Huggingface libraries~\cite{wolf2019huggingface}. All the binarization processes and experiments are conducted on a single 80 GB NVIDIA A100. Given that \textit{BiLLM} is an efficient PTQ framework, it eliminates the need for any fine-tuning, allowing for completion through a single quantization process.

\begin{table*}[!ht]
\centering
\setlength{\tabcolsep}{3.7mm}
\caption{
Perplexity of RTN, GPTQ, PB-LLM, BiLLM on LLaMA Family. The columns represent the perplexity results on Wikitext2 datasets with different model sizes. 
}
\vspace{-0.1in}
\begin{tabular}{llcrrrrr}
\toprule
\textbf{Model} & \textbf{Method} & \makecell{\textbf{Block} \\ \textbf{Size}} & \makecell{\textbf{Weight} \\ \textbf{Bits}}  & \textbf{7B} & \textbf{13B} & \textbf{30B} & \textbf{65B/70B*} \\
\midrule
 & Full Precision  & - & 16.00 & 5.68 & 5.09 & 4.10 & 3.53\\
 \cmidrule{2-8}
 & RTN & - & 2.00  & 106767.34 & 57409.93 & 26704.36 & 19832.87\\
 & GPTQ & 128 &  2.00  & 152.31 & 20.44 & 13.01 & 8.78\\
\textbf{LLaMA} & RTN & - & 1.00  & 168388.00 & 1412020.25 & 14681.76 & 65253.24\\
 & GPTQ & 128 & 1.00 & 267001.72 & 113894.12 & 67093.73 & 25082.88\\
 & PB-LLM $\dagger$ & 128 & 1.70  & 102.36 & 36.60 & 33.67 & 12.53\\
 & \textbf{BiLLM} $\ddagger$ & 128 & \textbf{1.09}  & \textbf{35.04} & \textbf{15.14} & \textbf{10.52} & \textbf{8.49}\\
\cmidrule{1-8}
 & Full Precision & - & 16.00 & 5.47 & 4.88 & N/A & 3.32\\
 \cmidrule{2-8}
 & RTN & - & 2.00 & 17788.93 & 51145.61 & N/A & 26066.13\\
 & GPTQ & 128 & 2.00  & 60.45 & 19.70 & N/A & 9.12\\
 \textbf{LLaMA2} & RTN & - & 1.00  & 157058.34 & 47902.32 & N/A & 160389.91\\
 & GPTQ  & 128 & 1.00 & 115905.67 & 9387.80 & N/A & 74395.42\\
 & PB-LLM $\dagger$ & 128 & 1.70  & 69.20 & 151.09 & N/A & 28.37\\
 & \textbf{BiLLM} $\ddagger$ & 128 & \textbf{1.08}  & \textbf{32.48} & \textbf{16.77} & N/A & \textbf{8.41}\\
\bottomrule

\end{tabular}
\begin{tablenotes}
        \footnotesize
        \item  The table gives the average bit-width of the LLaMA family. $\text {N/A}$: LLaMA2 do not have 30B version. *: LLaMA has 65B version and LLaMA2 has 70B version.
\end{tablenotes}
\label{exp_llama}
\end{table*}

\textbf{Models and Datasets.}
We facilitate our method on the OPT~\cite{zhang2022opt} and LLaMA~\cite{touvron2023Llama,touvron2023Llama2} families. Additionally, considering the customary need for instruction-based fine-tuning of LLMs to adapt to varying contexts, we also conducted experiments on Vicuna~\cite{chiang2023vicuna}. In terms of evaluation metrics, we mainly focused on the perplexity of LLMs' outputs, which is widely acknowledged in prior studies as a challenging yet stable indicator of LLM capabilities, particularly apt for network compression ~\cite{yao2206efficient,frantar2022gptq,frantar2023sparsegpt,xiao2023smoothquant}. We consider the test of WikiText2~\cite{merity2016pointer}, PTB~\cite{marcus1994penn}, as well as a part of the C4~\cite{raffel2020exploring} data. Then, we further conduct the experiments on seven zero-shot evaluation tasks (PIQA~\cite{bisk2020piqa}, BoolQ~\cite{clark2019boolq}, OBQA~\cite{mihaylov2018can}, Winogrande~\cite{sakaguchi2021winogrande}, ARC-e~\cite{clark2018think}, ARC-c~\cite{clark2018think} Hellaswag~\cite{zellers2019hellaswag}) in the Appendix \ref{ap_more_exp}, further verifying the robustness of our proposed \textit{BiLLM} to the binarization of LLMs. 

\textbf{Baseline.}
Our primary baseline is PB-LLM~\cite{shang2023pb}, the most recent PTQ approach on binary LLMs. GPTQ~\cite{frantar2022gptq} and vanilla RTN are also selected. GPTQ is currently the advanced technology in PTQ, and many works\cite{lin2023awq,dettmers2023spqr,shang2023pb} choose it as the baseline. Other methods oriented towards 8-bit and 4-bit quantization are deemed unsuitable for binarization and were thus not considered. 

\subsection{Results}

\textbf{Comparison results.}
We conduct a meticulous comparison of the binary performance of different LLMs across various model sizes. We deploy the \textit{BiLLM} on the OPT models~\cite{zhang2022opt} under the condition of a block size equal to 128. As seen in Table \ref{exp_opt}, the model outputs under the RTN and GPTQ methods have already collapsed at 1-bit weights, whereas \textit{BiLLM} still maintains reasonable linguistic output capabilities with an average weight of \textbf{1.1} bits. In comparison with PB-LLM at 1.7 bits, our method achieves a 35\% reduction in weight bit-width while enhancing the performance of different sizes of the OPT model by 49.4\% to 77.0\%. It is noteworthy that when the parameter size exceeds 30B, \textit{BiLLM} can achieve performance nearly equivalent to that of GPTQ with 3-bit quantization.
\begin{figure*}[!t]
\centerline{\includegraphics[width=1.\textwidth]{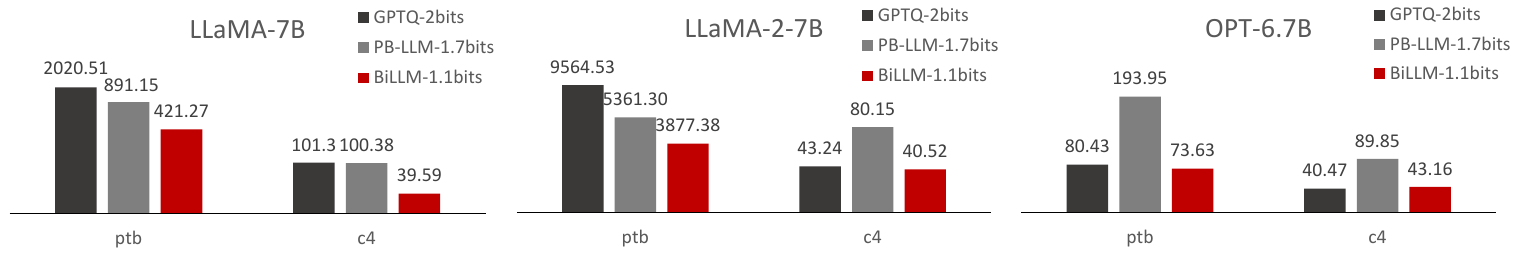}}
\vspace{-0.15in}
\caption{GTPQ, PB-LLM, BiLLM performed on the PTB and c4 datasets, mainly on LLaMA-7B, LLaMA2-7B, and OPT-6.7B, and we found that BiLLM performed relatively well.}
\label{ptb_c4}
\end{figure*}
\begin{table}[!t]
\centering
\setlength{\tabcolsep}{.7mm}
\caption{
Perplexity of \textit{BiLLM} on Vicuna-7B and Vicuna-13B. The columns of different models represent the perplexity results on Wikitext2, PTB, and C4 datasets. The block size is set to 128.
}
\vspace{-0.1in}
\begin{tabular}{llccrr}
\toprule
\textbf{Model} & \textbf{Method} & \makecell{\textbf{Weight} \\ \textbf{Bits}} & \makecell{\textbf{Wiki}\\ \textbf{-text2}} $\downarrow$ & \textbf{PTB} $\downarrow$ & \textbf{C4} $\downarrow$\\

\midrule

 & GPTQ & 2.00 & 109.56 & 6227.73 & 64.28\\
\textbf{Vicuna-7B} & PB-LLM & 1.70 & 68.01 & 477.52 & 67.23\\
 & \textbf{BiLLM} & \textbf{1.08} & \textbf{33.00} & \textbf{332.17} & \textbf{36.24}\\
\cline{1-6}
 & GPTQ & 2.00 & 41.75 & 465.94 & 40.57\\
\textbf{Vicuna-13B} & PB-LLM & 1.70 & 362.17 & 772.44 & 346.16\\
 & \textbf{BiLLM}& \textbf{1.08} & \textbf{36.57} & \textbf{300.31} & \textbf{28.76}\\
\bottomrule

\end{tabular}

\label{result_vicuna}
\end{table}

Due to the exceptional performance of the LLaMA~\cite{touvron2023Llama, touvron2023Llama2} series, they have become the foundation for many open-source models~\cite{chiang2023vicuna}. Then, in Table \ref{exp_llama}, we evaluate the perplexity of outputs from the LLaMA series models using different methods. It can be observed that, even at ultra-low weight bit-width, \textit{BiLLM} consistently outperforms the 2-bit RTN and GPTQ methods. And \textbf{1.08} bits \textit{BiLLM} for LLaMA-65B and LLaMA2-70B even surpasses the output of the full-precision OPT-66B model, which demonstrates the further binary potential of the LLaMA family. We extend perplexity evaluation to the PTB and C4 datasets. Figure~\ref{ptb_c4} illustrates the performance of the 7B parameter LLaMA series as well as the 6.7B OPT models. \textit{BiLLM} continues to achieve a leading edge in performance compared to other methods (more additional comparisons are discussed in Appendix \ref{ap_more_exp}).

\begin{figure}[!t]
\vspace{-0.1in}
\centerline{\includegraphics[width=1.\columnwidth]{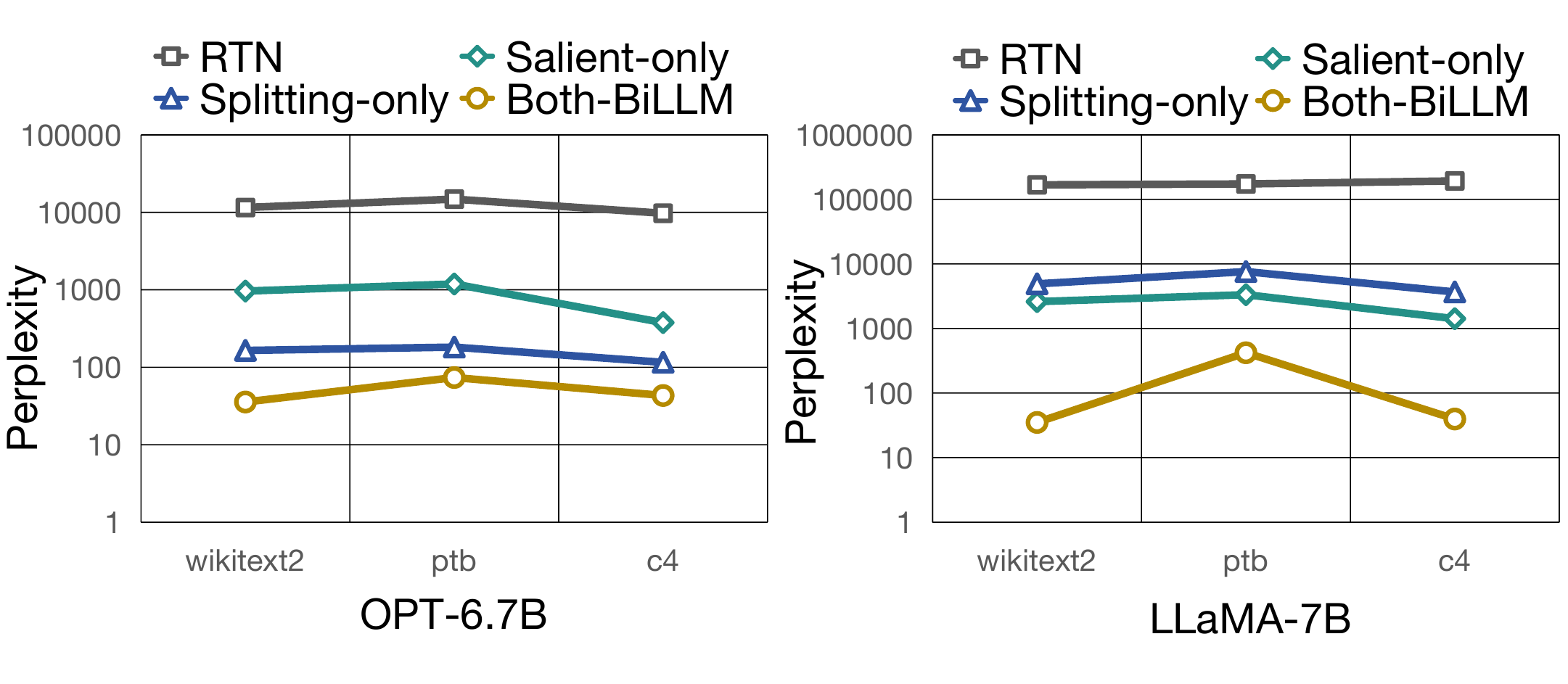}}
\vspace{-0.2in}
\caption{Ablation results of salient-only and splitting-only methods on OPT and LLaMA. }
\label{ablation_method}
\vspace{-0.1in}
\end{figure}

\begin{table*}[!h]
\centering
\setlength{\tabcolsep}{1.5mm}
\caption{
Model size comparison of LLaMA family.
}
\vspace{-0.1in}
\begin{tabular}{lrrrrrrr}
\toprule
\textbf{Method} & \textbf{LLaMA-7B} & \textbf{LLaMA2-7B} & \textbf{LLaMA-13B} & \textbf{LLaMA2-13B} & \textbf{LLaMA-30B} & \textbf{LLaMA-65B} & \textbf{LLaMA-70B} \\ \midrule
FP16 & 13.5GB & 13.5 GB & 24.2 GB & 25.0 GB & 60.5 GB & 121.0 GB & 129.3 GB \\ 
BiLLM & 1.5 GB & 1.6 GB & 2.7 GB & 2.8 GB & 6.1 GB & 14.8 GB & 15.4 GB \\
\bottomrule
\end{tabular}
\label{billm_storage}
\end{table*}

\textbf{Experiments of instruction-tuned models.}
Instruction fine-tuning can significantly improve the application capabilities of the model and has become a necessary process for LLMs deployment in different scenarios~\cite{wei2021finetuned,sanh2021multitask,chiang2023vicuna}. We also deployed \textit{BiLLM} on the recently popular fine-tuning instruction model Vicuna for benchmark testing. As shown in Table \ref{result_vicuna}, the perplexity performance of GPTQ and PB-LLM are compared on Vicuna-7B and Vicuna-13B with three evaluations. \textit{BiLLM} can achieve better performance at an average weight bit of \textbf{1.08}, which further proves that \textit{BiLLM}’s universal LLMs binarization potential. We also provide dialogue examples of binary models in Appeandix \ref{ap_dialog}.

\textbf{Zero-Shot results.}
To conduct a more comprehensive evaluation of binary LLMs, we extend our experiments to 7 zero-shot datasets. Appendix \ref{ap_more_exp} provides detailed results of our approach compared to previous methods in ultra-low bit quantization, further showing the outlier of \textit{BiLLM}.

\textbf{Ablation results.}
\textit{BiLLM} enhances binarization precision through two primary methods: structured salient binarization via residual approximation, and non-salient weight binarization via optimal splitting. To examine the effects of these strategies, we conducted decomposition experiments. As shown in Figure~\ref{ablation_method}, both approaches significantly improve binary performance. Notably, we found that OPT-6.7B exhibits greater sensitivity to the splitting of non-salient weights (the blue line is lower than the green line), whereas LLaMA-7B is more responsive to salient weights' residual approximation (the green line is lower than the blue line). This further indicates that different LLMs exhibit varying responses to distinct binarization optimization strategies, showing that the two binarization strategies proposed by \textit{BiLLM} are efficient to various LLMs. We further discuss details on the block-size ablation results in Appendix \ref{ap_block}.

\begin{table}[!t]
\vspace{-0.1in}
\centering
\setlength{\tabcolsep}{0.8mm}
\caption{
The memory occupancy rate compared with FP16 and the corresponding accuracy on OPT-30B.
}
\vspace{-0.1in}
\begin{tabular}{lrrr}
\toprule
\textbf{Configuration} & \textbf{BiLLM} & \textbf{PB-LLM (10\%)} & \textbf{GPTQ} \\ \midrule
Average bit-width & 1.11 & 1.7 & 2 \\ \midrule
Memory Occupancy$^*$ & 9.70\% & 16.50\% & 13.30\% \\ \midrule
PPL on WikiText-2 & 12.71 & 25.14 & 15.71 \\ 
\bottomrule
\end{tabular}
\begin{tablenotes}
\footnotesize
\item *: Equation of memory occupancy [R6]: memory\_occupancy\_compared\_w\_FP16 = (binary\_unsalint\_weight\_size + residual\_binary\_salint\_weight\_size + CSR\_compressed\_bitmap\_size + scaling\_factor\_size) / floating\_point\_weight\_size.
\end{tablenotes}
\label{billm_gpu_1}
\end{table}

\begin{table}[!t]
\centering
\setlength{\tabcolsep}{6.6mm}
\caption{
Memory occupancy rate compared with FP16 OPT.
}
\vspace{-0.1in}
\begin{tabular}{lrr}
\toprule
\textbf{Model} & \textbf{BiLLM} & \textbf{GPTQ-2bit} \\ \midrule
OPT-1.3B & 9.40\% & 13.30\% \\ 
OPT-2.7B & 9.50\% & 13.30\% \\ 
OPT-6.7B & 9.30\% & 13.30\% \\ 
OPT-13B & 9.70\% & 13.30\% \\ 
OPT-30B & 9.70\% & 13.30\% \\ 
\bottomrule
\end{tabular}
\vspace{-0.1in}
\label{billm_gpu_2}
\end{table}

\textbf{Model size.}
In Table~\ref{billm_storage}, we present the FP16 of models ranging from LLaMA-7B to 65B and LLaMA2-7B to 70B, as well as the sizes of binarized models after compression by BiLLM. Notably, BiLLM achieved close to a tenfold compression of weights across LLMs of different sizes.

\textbf{GPU memory.}
The motivation of our BiLLM is to push the bit-width compression limit of LLM weights under post-training conditions, which reduces both the storage and GPU memory footprint of LLMs and retains their accuracy to the greatest extent for being practical. Although binarized GEMM is hard to implement directly due to fine-grained grouping, the extreme bit-width compression of our BiLLM brings significant savings in GPU memory requirements (size and bandwidth), which is considered to be one of the most significant efficiency bottlenecks of LLM inference~\cite{gholami2024ai,dettmers2208llm,dettmers2023qlora,xiao2023smoothquant,chee2024quip,shang2023pb}. Here, we provide detailed memory and performance comparisons to demonstrate the advantages of BiLLM (as shown in Table A.1): for the OPT-30B model, BiLLM (1.1-bit) achieves a 41.57\% and 27.07\% memory compression improvement compared to PB-LLM (1.7-bit) and GPTQ (2-bit), respectively, while enhancing accuracy by 49.44\% and 19.10\%. 
We further provide a detailed comparison of memory usage with the 2-bit GPTQ method under different sizes of LLM in Table~\ref{billm_gpu_2}. The memory occupancy of our BiLLM is only about 69.9\% of 2-bit quantization, which shows the great memory-saving benefit of our BiLLM from the extreme bit-width reduction, and we also achieve higher accuracy with the significantly saved memory.

\section{Conclusions}
This work proposed a novel post-training binary quantization method named \textit{BiLLM}, specifically tailored for compressing pre-trained LLMs. Inspired by the characteristics of weight's value and Hessian distributions, we adopted a binary residual approximation for structurally salient weights to preserve their capabilities at ultra-low bits. For non-salient weights, we employed optimal segmentation for grouped binarization. Our results demonstrate that LLMs can undergo a one-time weight quantization at ultra-low bits without substantial loss of precision. \textit{BiLLM} has pioneered the achievement of LLM performance guarantees at an average bit rate close to 1 bit. We validated the binarization performance of \textit{BiLLM} across multiple open-source LLM families and conducted generalization tests on a fine-tuned instruction model. \textit{BiLLM} advances the bit-width quantization frontier of LLMs, promising to facilitate the deployment of LLMs in edge scenarios and resource-constrained devices, and encourages further exploration in LLMs compression. 

\textbf{Acknowledgement.} This work was supported by the National Science and Technology Major Project (2021ZD0110503), the Swiss National Science Foundation (SNSF) project 200021E\_219943 Neuromorphic Attention Models for Event Data (NAMED), the Baidu Scholarship, and the National Natural Science Foundation of China (No. 62306025, No. 92367204).

\nocite{langley00}

\section*{Impact Statement}

This paper presents work whose goal is to advance the field of Machine Learning. There are many potential societal consequences of our work, none which we feel must be specifically highlighted here.

{
\small
\bibliography{example_paper}
\bibliographystyle{icml2024}
}

\newpage
\appendix
\onecolumn

\section {\textit{BiLLM} Implementation}\label{func}
\begin{algorithm*}[!h]
\begin{multicols}{2}
\caption{BiLLM: Detailed functions process}
\label{alg2}
func $\operatorname{salient}$ $(\mathbf{W}, \mathbf{H^c})$
\begin{algorithmic}[1]

\STATE $\mathbf{S} \coloneqq  \mathbf{W}^2 / [\mathbf{H}^c_{b:b+\beta b:b+\beta}]^2 \ $ // \ salient matrix
\STATE $row_s\{\cdot\} \coloneqq \ \operatorname{topk} (\operatorname{sum}(\operatorname{abs}(\mathbf{S})).(\dim=0)) $
\STATE $e = \operatorname{inf}$ // \ searching error
\STATE ${n}^* = 0$ // \ optimal number of salient columns
\FOR{$i=1,2,..., \operatorname{len}(row_s)$}
    \STATE$\mathbf{B}_1 \coloneqq \ \operatorname{binary}(\mathbf{W}_{{:,j}, \ j \in row_s[:i]})$
    \STATE$\mathbf{B}_2 \coloneqq \ \operatorname{binary}(\mathbf{W}_{{:,j}, \ j \notin row_s[:i]})$
    \IF{$||\mathbf{W} - (\mathbf{B}_1 \cup \mathbf{B}_2)||^2 < e$}
        \STATE$e \coloneqq ||\mathbf{W} - (\mathbf{B}_1 \cup \mathbf{B}_2)||^2$
        \STATE${n}^* \coloneqq  i$
    \ENDIF
\ENDFOR
\STATE {\bf return}  $row_s\{ : {n}^* \}$
\end{algorithmic}

func $\operatorname{binary}$ $(\mathbf{W})$
\begin{algorithmic}[1]
\STATE $\alpha \coloneqq \dfrac{||\mathbf{W}||_{\ell1}}{m}$
\STATE $\mathbf{B} \coloneqq \alpha \cdot \operatorname{sign}(\mathbf{W})$
\STATE {\bf return}  $\mathbf{B}$
\end{algorithmic}

func $\operatorname{res\_approximation}$ $(\mathbf{W})$
\begin{algorithmic}[1]
\STATE $\mathbf{B}_1 \coloneqq \operatorname{binary} (\mathbf{W})$
\STATE $\mathbf{R} \coloneqq \mathbf{W} - \mathbf{B}_1$
\STATE $\mathbf{B}_2 \coloneqq \operatorname{binary} (\mathbf{R})$
\STATE $\mathbf{B} \coloneqq \mathbf{B}_1 + \mathbf{B}_2$
\STATE {\bf return}  $\mathbf{B}$
\end{algorithmic}

func $\operatorname{seg\_search}$ $(\mathbf{W})$
\begin{algorithmic}[1]
\STATE $e = \operatorname{inf}$ // \ searching error
\STATE $p^* = 0$ // \ optimal break-point
\FOR{$i=0.1, 0.2, 0.3,...,9$}
    \STATE$p \coloneqq  i \cdot \operatorname{max}(\operatorname{abs}(\mathbf{W}))$
    \STATE$\mathbf{B}_1 \coloneqq \ \operatorname{binary} (\mathbf{W}_{|w_{i,j}| \leq p})$
    \STATE$\mathbf{B}_2 \coloneqq \ \operatorname{binary}(\mathbf{W}_{|w_{i,j}| > p})$
    \IF{$||\mathbf{W} - (\mathbf{B}_1 + \mathbf{B}_2)||^2 < e$}
        \STATE$e \coloneqq ||\mathbf{W} - (\mathbf{B}_1 + \mathbf{B}_2)||^2$
        \STATE$p^* \coloneqq  p$
    \ENDIF
\ENDFOR
\STATE {\bf return}  $p^*$
\end{algorithmic}

\end{multicols}
\end{algorithm*}

\textit{BiLLM} necessitates the structured selection of salient rows and their subsequent quantization through residual approximation binarization. This is followed by dividing the non-salient weights, which exhibit a bell-shaped distribution, into a sparse area and a concentrated area. The division requires the optimization of the segmentation point $p^*$ by minimizing quantization loss. Ultimately, the two regions of non-salient weights are binarized separately to derive the final binary weights for LLMs. The implementation details of the aforementioned function are enumerated in Algorithm \ref{alg2}.
\section {Quantization Error} \label{ap_msqe}
\textbf{Quantization error definition for weight distribution}
The numerical range covered by the uniform quantizer spans from $ [X_{min}, X_{max}]$. The number of intervals post-quantization, denoted as $M$, typically equals $2^b$, where $b$ represents the target bit-width of quantization. So the quantization step size is: 
\begin{equation} \label{apeq1}
    \Delta = \frac{X_{\text{max}} - X_{\text{min}}}{M}
\end{equation}
The boundaries can be calculated as:
\begin{equation} \label{apeq2}
    b_q = X_{\text{min}} + \Delta \cdot l
\end{equation}
where $l \in {0,1,...,M}$, and we have $b_q \in \{-\alpha, 0, \alpha\}$ under binarization. Then we give the mean of each interval:
\begin{equation} \label{apeq3}
    x_q = X_{\text{min}} + \Delta \cdot l - 0.5\Delta
\end{equation}
where $l \in {1,...,M}$. In this quantization scheme, we can get the MSQE from ~\cite{you2010audio}:
\begin{equation} \label{apeq4}
    \theta^2 = \sum_{l=1}^{M} \int_{X_{\text{min}} + \Delta \cdot (l-1)}^{{X_{\text{min}} + \Delta \cdot l}} (X_{\text{min}} + \Delta \cdot l - 0.5\Delta - x)^2 g(x) dx
\end{equation}
then we let the $y$ to replace the $X_{\text{min}} + \Delta \cdot l - 0.5\Delta - x$ part, so the Equation~\eqref{apeq4} becomes: 
\begin{equation} \label{apeq5}
    \theta^2 = \sum_{l=1}^{M} \int_{-0.5\Delta}^{0.5\Delta} y^2f[X_{\text{min}} + \Delta \cdot l - (y + 0.5\Delta)]^2 dx
\end{equation}
consider the Equation~\eqref{apeq2} and Equation~\eqref{apeq3}, the above equation becomes:
\begin{equation} \label{apeq6}
    \theta^2 = \sum_{l=1}^{M} \int_{-0.5\Delta}^{0.5\Delta} x^2f(x_p - x) dx
\end{equation}
The aforementioned reasoning indicates that the MSQE of a uniform quantizer depends on the PDF and the quantization bit-width. Due to previous observations of the weights in pretrained LLMs, we have eliminated the salient weights. The remaining distribution of non-salient weights' $g(x)$, is not uniform and resembles a Gaussian distribution. In binarization, therefore, we substitute $\alpha$ into Equation~\eqref{apeq4}, resulting in:
\begin{eqnarray}    \label{eq}
\theta^2 &=& \sum_{l=1}^{M} \int_{(l-1-0.5M)\Delta}^{(l-0.5M)\Delta} [(l-0.5-0.5M)\Delta - x]^2 g(x) dx  \nonumber    \\
&=&\int_{X_{\text{min}}}^{0} (-\alpha - x)^2 g(x) dx + \int_{0}^{X_{\text{max}}} (\alpha - x)^2 g(x) dx
\end{eqnarray}

\section {Searching Curve of Salient Column and Non-salient Distribution} \label{ap_seg}
\begin{figure}[!h]
\centerline{\includegraphics[width=1\columnwidth]{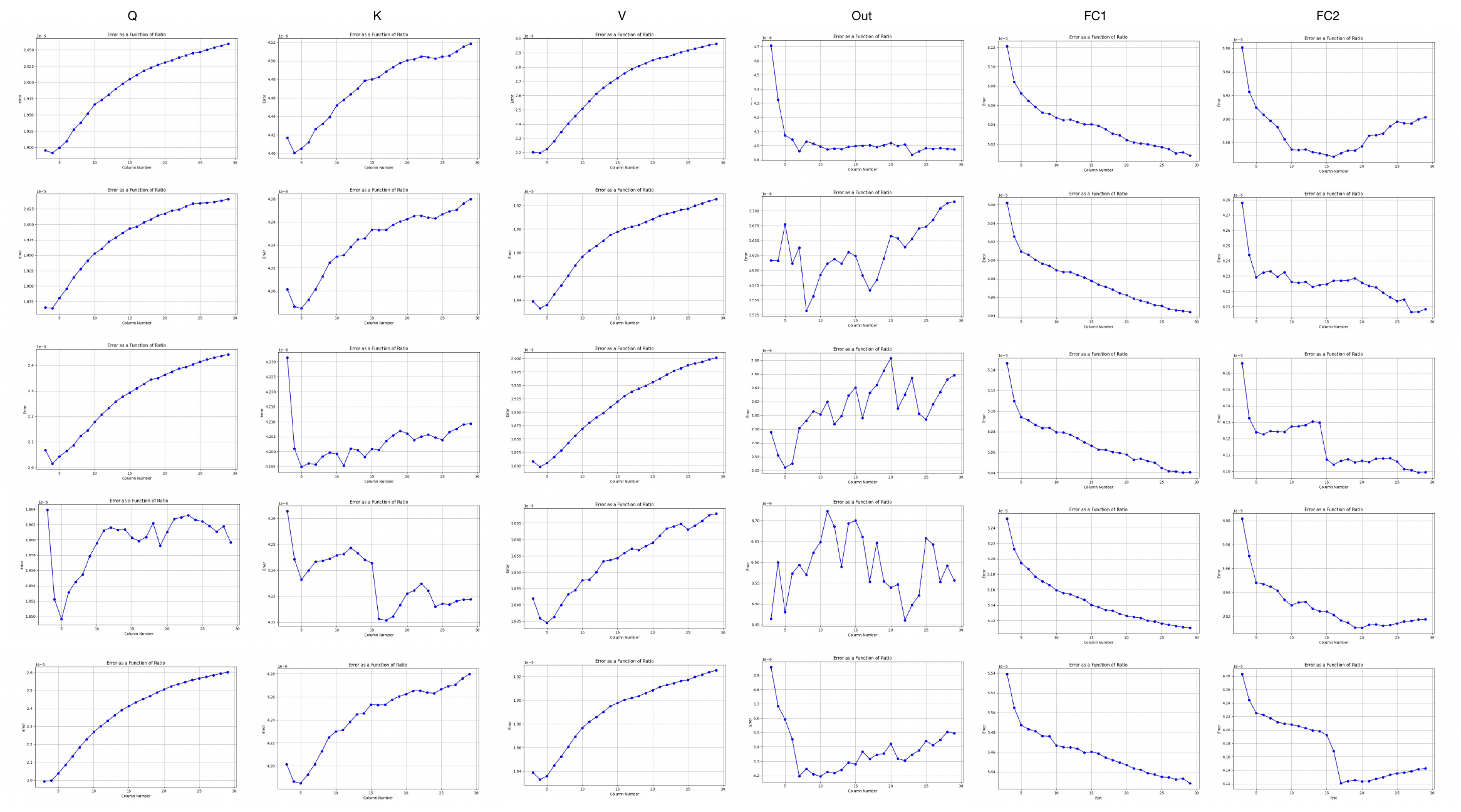}}
\caption{Block-wise searching curve of salient columns in OPT-6.7B. The majority of the curves indicate that the minimal quantization error can be achieved at the block level by considering only a few columns as salient. The \textit{Out Projection} layer has a larger number of salient columns, hence varying coverage for each block. The distribution in the \textit{FC} layer is more dispersed. After optimal searching, the overall average weight bit is merely \textbf{1.1} bits.}
\label{ap_search_structure}
\end{figure}
We implemented a column-level segmentation and formulated a minimal-error column number search, as delineated in Equation~\eqref{eq_column_search}. The identification of the optimal count of salient column groups commences with the column exhibiting the highest salience. To mitigate the increase in bit-width resulting from residual approximation, we confined the search range to between 3 to 30 columns. Figure~\ref{ap_search_structure} illustrates the search curve pertinent to the inaugural Transformer block within the OPT6.7B model. It includes six layers of operators (\textit{Q}, \textit{K}, \textit{V}, \textit{Out Projection}, \textit{FC1}, and \textit{FC2}), with each layer showing the search curves for the first five blocks. Figure~\ref{ap_opt_hessian} elucidates the clustering of salient weights, suggesting that a majority of the layers and blocks are capable of attaining minimal quantization errors with a limited number of salient columns. The block-wise changes in weight distribution brought about by OBC~\cite{frantar2022optimal} introduce fluctuations in the search curve; however, the structured selection still manages to encompass the majority of salient weights. In the \textit{Feedforward} layer, where salient weight distribution is more scattered, the search curve leans towards employing residual approximation across an increased number of columns. Nonetheless, Table \ref{salient_searching_bits}, displaying the average weight bit numbers across various LLMs, confirms that this search strategy effectively maintains weight compression at approximately \textbf{1.1} bits.

\begin{figure}[!h]
\centerline{\includegraphics[width=1\columnwidth]{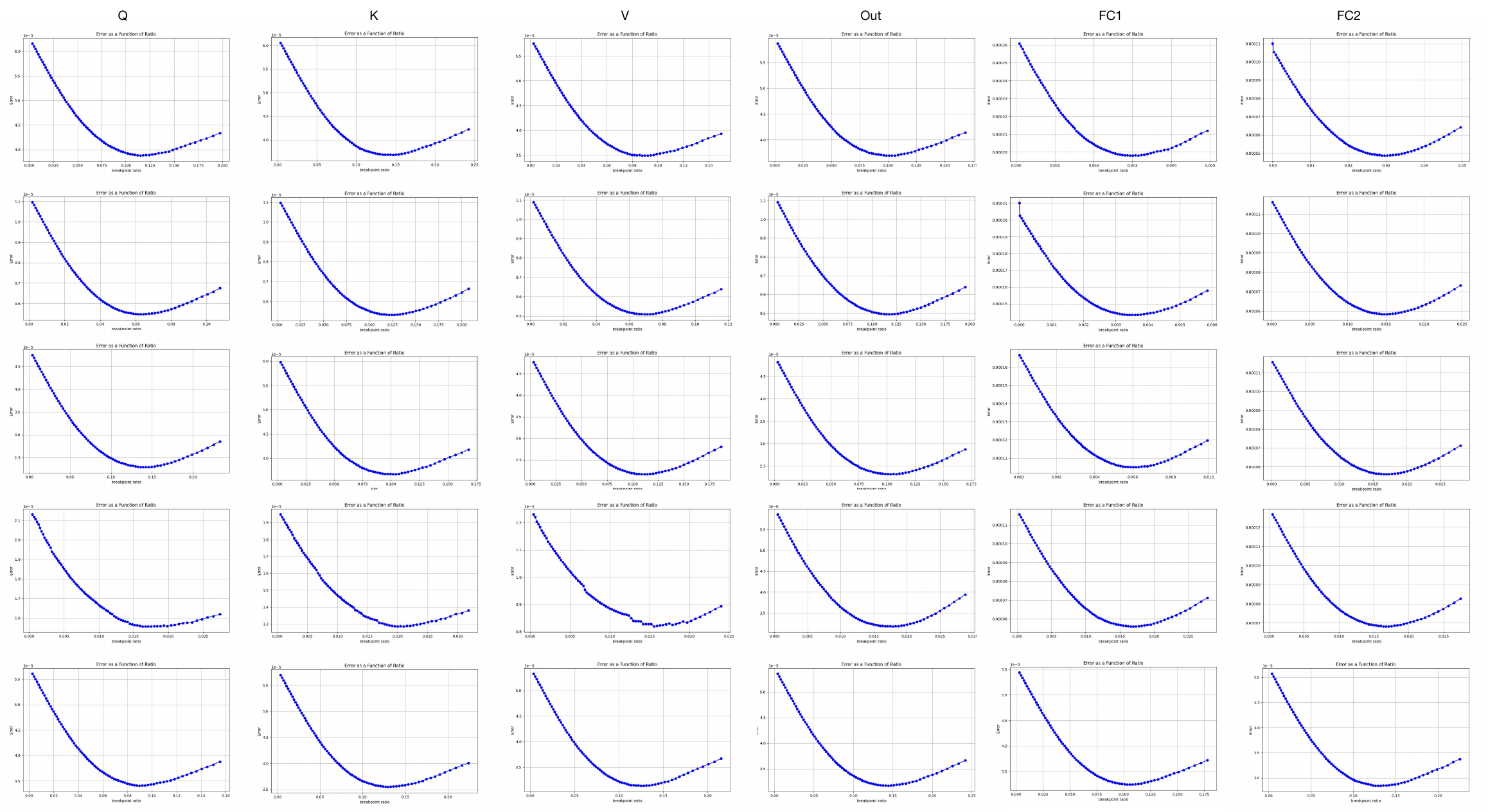}}
\caption{Block-wise splitting curve of bell-shaped distribution in OPT6.7B. The overall presentation exhibits the characteristics of a convex function, fundamentally aligning with the theoretical optimal point in terms of theoretical basis.}
\label{ap_search_distribution}
\end{figure}

Figure~\ref{ap_search_distribution} shows the unstructured search curve for the non-salient weights in the OPT6.7B model, with the same composition as that in  Figure~\ref{ap_search_structure}. The horizontal axis represents the ratio between  $p$ and the maximum weight value. Despite searching on a block-wise basis, the search curve still exhibits convex properties, indicating the presence of an optimal $p*$. This phenomenon demonstrates that the non-salient weights exhibit characteristics closely resembling an ideal Gaussian or Laplacian distribution~\cite{you2010audio,fang2020post}.

\section{Multi-evaluation Comparisons} \label{ap_more_exp}

\begin{figure*}[!h]
\centerline{\includegraphics[width=1.\textwidth]{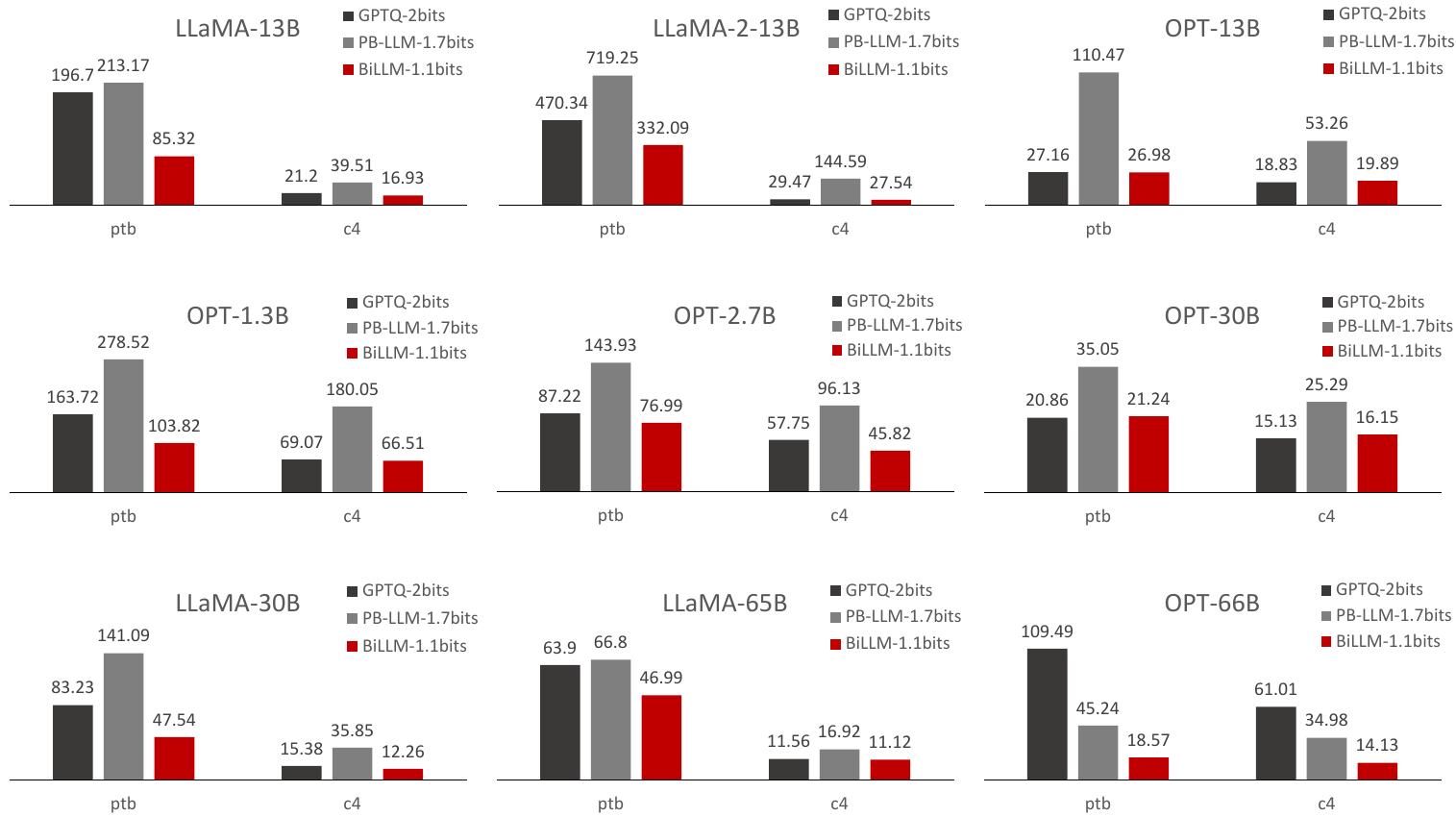}}
\caption{GPTQ, PB-LLM, BiLLM performed on the PTB and C4 datasets, mainly on LLaMA-13B, LLaMA2-13B, OPT-13B, and so on. The results showed that BiLLM performed relatively well.}
\label{ptb_c4_2}
\end{figure*}

\textbf{Perplexity results on PTB and C4.}

We use tables in the main text to show the perplexity of the three methods GPTQ, PB-LLM, and BiLLM on the Wikitext2 dataset, and bar charts to show the perplexity results for LLaMA-7B, LLaMA2-7B, and OPT-6.7B on the PTB and C4 datasets. In the appendix, we show the quantitative comparison results for models of other sizes on the PTB and C4 datasets with more images.

In Figure~\ref{ptb_c4_2}, we find that although different models have different perplexity results, they still roughly follow the law that the larger the model, the lower the perplexity. BiLLM is generally still relatively better than the GPTQ and PB-LLM results in terms of perplexity with a lower bit-width configuration, while PB-LLM and GPTQ are higher or lower than each other, with slightly inferior results at very low bits.

\textbf{Zero-shot results}

For completeness of testing, we have also tested and compared metrics such as the accuracy of GPTQ, PB-LLM, and BiLLM on datasets such as PIQA and BoolQ, all using Zero Shot's experimental setup. From Table \ref{zeroshot_results}, We find that despite the loss in quantification, a side-by-side comparison between the three methods still shows BiLLM to be superior overall, testing one level higher on some datasets, while the effect of some random perturbations, although present, does not pull down BiLLM's performance across the board. This suggests that BiLLM's quantization results have significantly improved performance at very low bits, and further validates the conclusions.

\begin{table*}[!ht]
\centering
\setlength{\tabcolsep}{0.8mm}
\caption{
Accuracy on 7 data sets, from binarization LLaMA, LLaMA2, and OPT, and we also compare the results among GPTQ, PB-LLM, and BiLLM to validate the quantization effect.
}
\begin{tabular}{llrcrrrrrrr}
\toprule
\textbf{Model} & \textbf{Method} & \makecell{\textbf{Weight} \\ \textbf{Bits}} & \makecell{\textbf{Block} \\ \textbf{Size}} & \textbf{PIQA} $\uparrow$& \textbf{BoolQ} $\uparrow$& \textbf{OBQA} $\uparrow$& \textbf{Winogrande} $\uparrow$& \textbf{ARC-e} $\uparrow$& \textbf{ARC-c} $\uparrow$ & \textbf{Hellaswag} $\uparrow$\\

\midrule

 & GPTQ & 2.00 & 128 & 52.8 & 50.0 & 28.2 & 49.3 & 26.6 & 29.5 & 26.3\\
\textbf{LLaMA-7B} & PB-LLM & 1.70 & 128 & 54.6 & 59.7 & 30.4 & 50.6 & 28.2 & 24.6 & 28.7\\
 & \textbf{BiLLM} & \textbf{1.09} & 128 & \textbf{61.2} & \textbf{62.7} & \textbf{31.8} & \textbf{51.1} & \textbf{36.0} & \textbf{25.7} & \textbf{36.8}\\
\midrule
 & GPTQ & 2.00 & 128 & 51.1 & 43.9 & 29.0 & 50.8 & 26.6 & 28.5 & 26.3\\
\textbf{LLaMA2-7B} & PB-LLM & 1.70 & 128 & 53.8 & 62.3 & 30.2 & 49.3 & 28.0 & 25.0 & 27.7\\
 & \textbf{BiLLM} & \textbf{1.08} & 128 & \textbf{60.6} & \textbf{61.8} & \textbf{33.2} & \textbf{52.4} & \textbf{36.2} & \textbf{24.4} & \textbf{34.8}\\
\midrule
 & GPTQ & 2.00 & 128 & 56.6 & 51.1 & 25.6 & 51.2 & 31.3 & 22.9 & 30.4\\
\textbf{OPT-6.7B} & PB-LLM & 1.70 & 128 & 57.6 & 55.5 & 24.2 & 47.7 & 33.2 & 21.0 & 31.0\\
 & \textbf{BiLLM} & \textbf{1.11} & 128 & \textbf{58.6} & \textbf{62.2} & \textbf{29.0} & \textbf{51.5} & \textbf{34.1} & \textbf{23.9} & \textbf{31.9}\\
 \bottomrule

\end{tabular}

\label{zeroshot_results}
\end{table*}

\section{Ablation of \textit{BiLLM} with different block size} \label{ap_block}

To explore the effect of different chunk sizes on the quantization effect of BiLLM, we set up block size settings including 32 columns and 64 columns up to 512 columns and performed quantization experiments on them. The results show that the overall perplexity is lower as the chunk granularity becomes finer and the number of bits used becomes relatively smaller. We believe this is because the smaller the chunks, the finer the data representation, and the more scale is used, but increasing the diversity of quantization results also increases the weighting overhead. A block size of 128 can better balance the bit-width and quantization effect.

\begin{table*}[!ht]
\centering
\setlength{\tabcolsep}{10mm}
\caption{
Perplexity on Wikitext2, PTB, and C4 with different block size settings on \textit{BiLLM}.
}
\begin{tabular}{lrrrrr}
\toprule
\textbf{Model} & \makecell{\textbf{Block} \textbf{Size}} & \textbf{Wikitext2} & \textbf{PTB} & \textbf{C4}\\

\midrule

 & 512 & 74.14 & 1078.90 & 81.76\\
 & 256 & 48.91 & 574.34 & 57.60\\
\textbf{LLaMA-7B} & \textbf{128} & \textbf{35.04} & \textbf{421.27} & \textbf{39.59}\\
 & 64 & 27.23 & 399.81 & 27.74\\
 & 32 & 17.56 & 263.39 & 19.85\\
\midrule

 & 512 & 52.90 & 267.82 & 43.86\\
 & 256 & 43.69 & 232.34 & 43.21\\
\textbf{LLaMA2-7B} & \textbf{128} & \textbf{32.48} & \textbf{3877.38} & \textbf{40.52}\\
 & 64 & 20.12 & 830.36 & 24.46\\
 & 32 & 13.58 & 440.40 & 17.34\\
\midrule

 & 512 & 151.81 & 257.22 & 101.96\\
 & 256 & 84.42 & 116.44 & 77.25\\
\textbf{OPT-6.7B} & \textbf{128} & \textbf{35.36} & \textbf{73.63} & \textbf{43.16}\\
 & 64 & 33.36 & 48.16 & 31.94\\
 & 32 & 20.48 & 31.02 & 21.47\\
\bottomrule

\end{tabular}

\label{diff_bs_results}
\end{table*}

\section{Dialog Examples} \label{ap_dialog}
In this section, we show some dialogue examples of binarized LLaMA-13B and Vicuna-13B. 
\begin{figure*}[!h]
\centerline{\includegraphics[width=1.\textwidth]{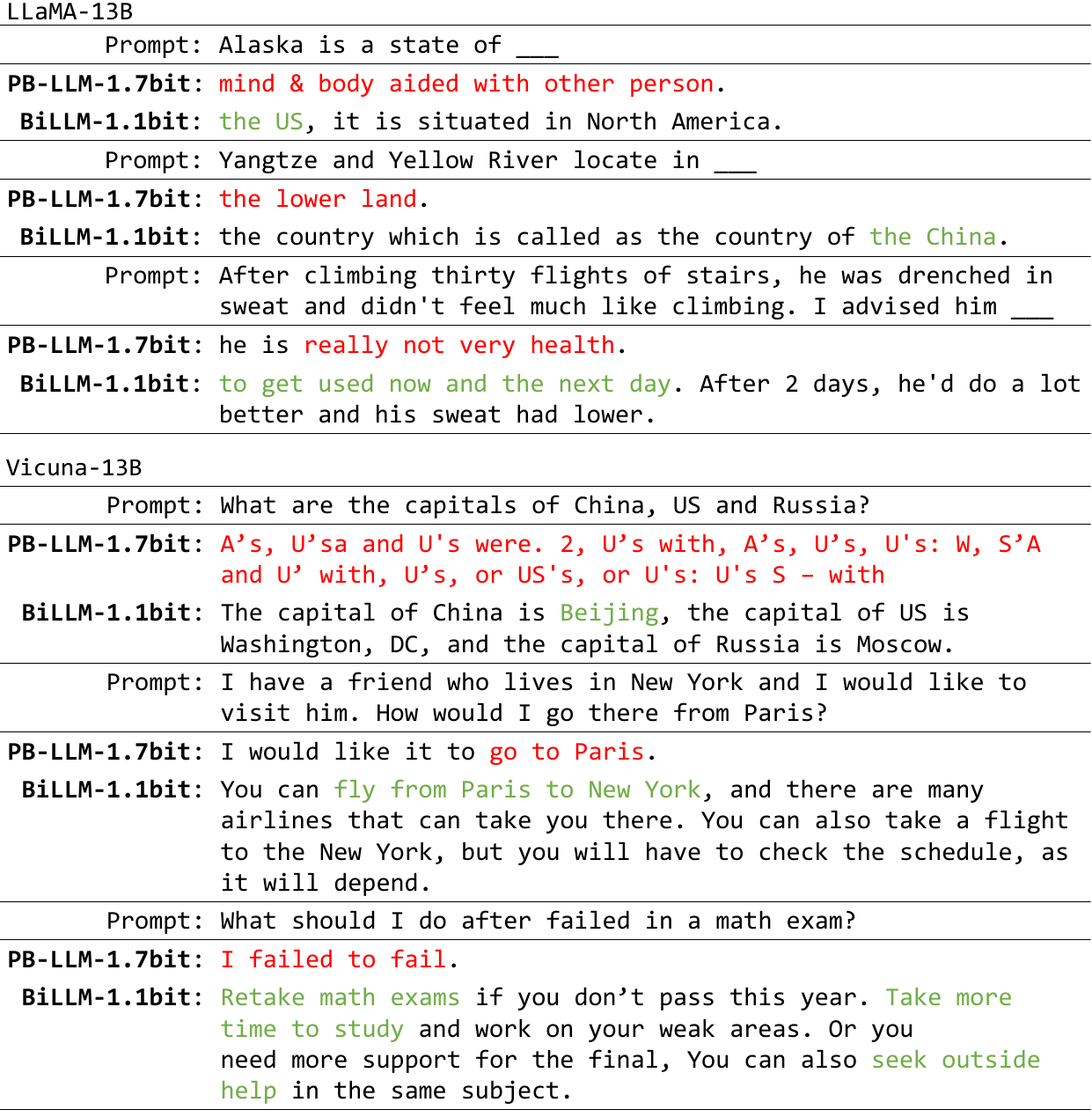}}
\caption{Some examples of conversations.  LLaMA-13B and Vicuna-13B are chosen to show the case of language supplementary and Q\&A ability. And PB-LLM (int 8, 10\%) is selected as the comparison. We color the text to show the \textcolor[rgb]{0.4375,0.6758,0.2773}{reasonable} or \textcolor[rgb]{1,0,0}{inappropriate} responses.}
\label{dialog}
\end{figure*}

\section {Magnitude and Hessian Distribution of LLMs}\label{more_dis}

\begin{figure}[!h]
\centerline{\includegraphics[width=1\columnwidth]{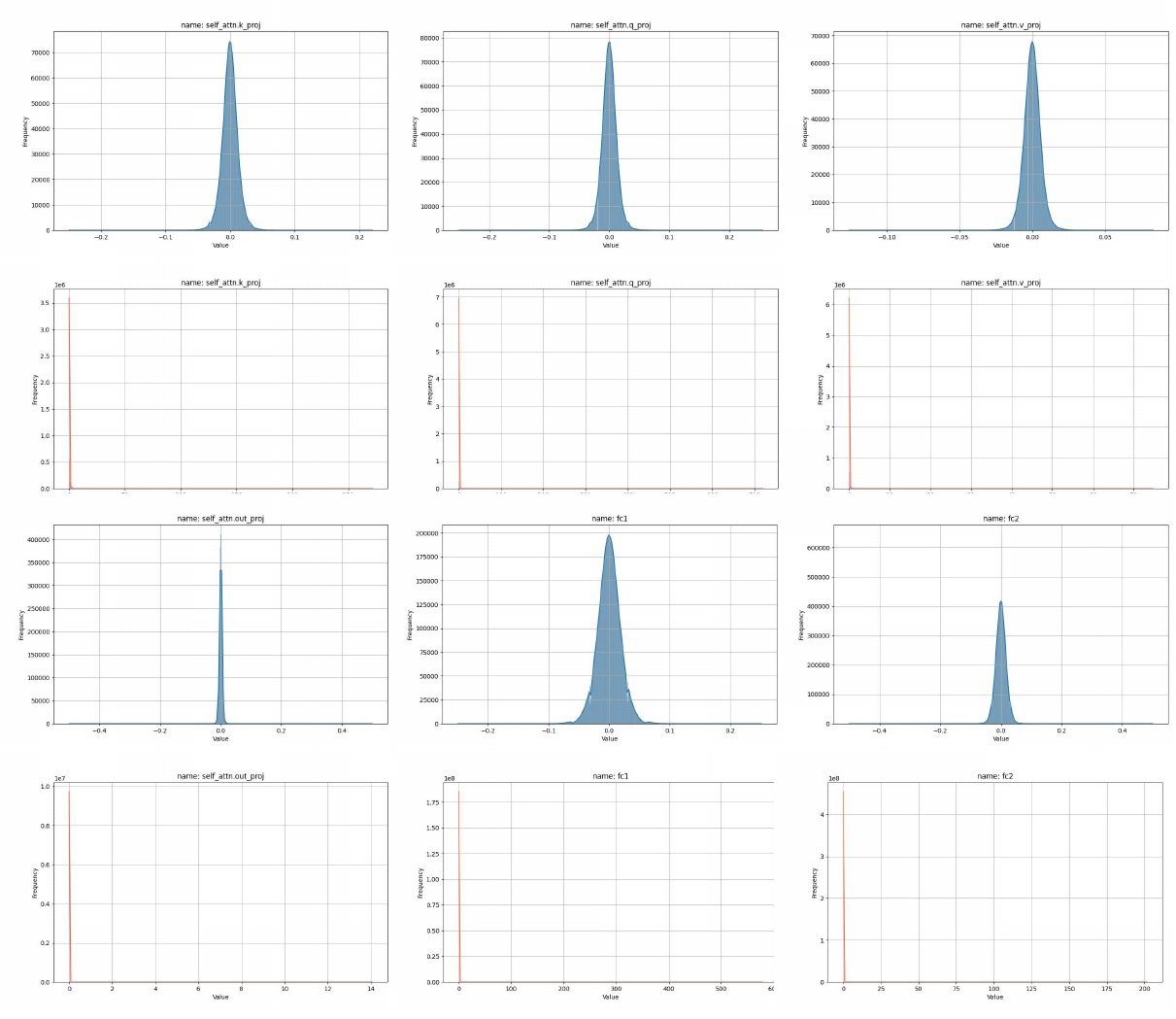}}
\caption{Different layers weight density distribution (blue) and hessian density distribution (orange) of the $1^{st}$ Transformer block of the OPT-1.3B model}
\label{ap_opt_dis}
\end{figure}

\begin{figure}[!h]
\centerline{\includegraphics[width=1\columnwidth]{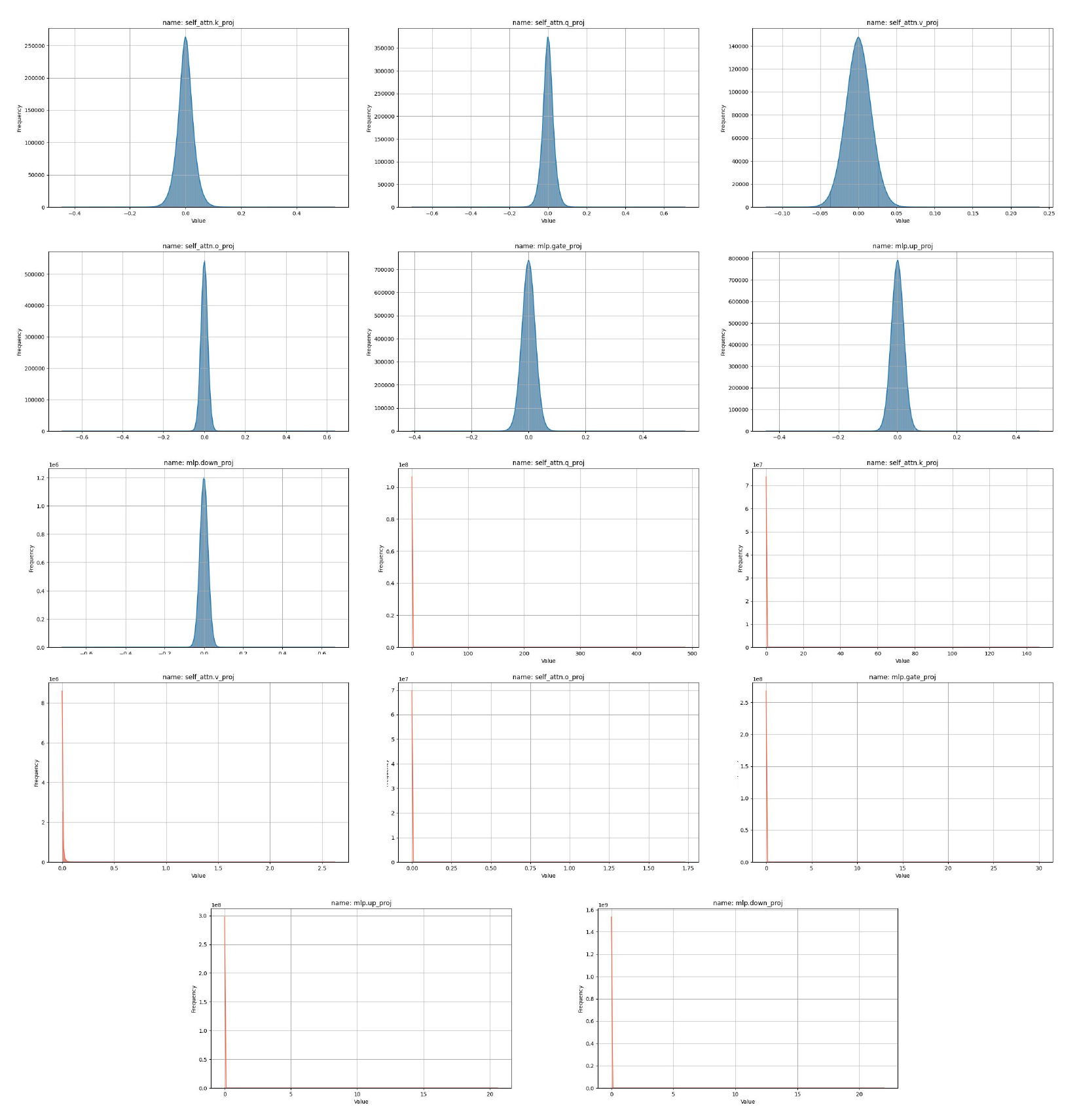}}
\caption{Different layers weight density distribution (blue) and hessian density distribution (orange) of the $6^{th}$ Transformer block of the LLaMA-7B model}
\label{ap_llama_dis}
\end{figure}

Figure~\ref{distribution} displays the distribution characteristics of weights and Hessian in LLMs. In this section, we provide additional examples to illustrate the bell-shaped distribution of weight values and the long-tailed distribution of Hessian weights. Figure~\ref{ap_opt_dis} depicts the distributions of four linear layers in the first Transformer block of the OPT-1.3B model, while Figure~\ref{ap_llama_dis} shows the distributions of seven linear layers in the sixth block of the LLaMA-7B model. The selection of these specific block positions is intended to demonstrate the universality of these distribution characteristics in LLMs.

\begin{figure}[!h]
\centerline{\includegraphics[width=1\columnwidth]{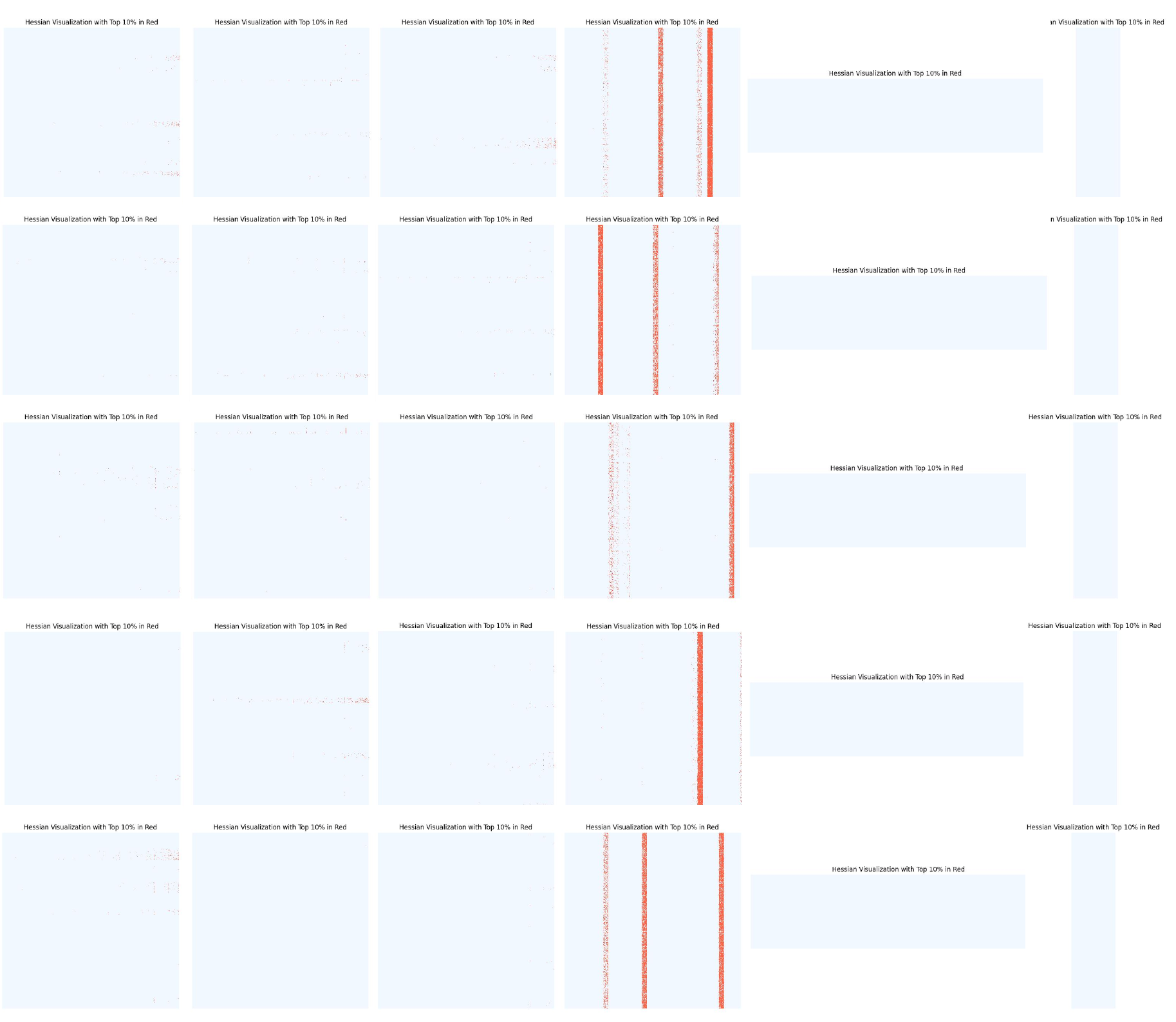}}
\caption{Distribution of top 10\% salient elements in Hessian matrix. The distribution of $1^{st}-5^{th}$ Transformer blocks in OPT-1.3B}
\label{ap_opt_hessian}
\end{figure}

Figure~\ref{ap_opt_hessian} displays the distribution of sensitive weights across 5 Transformer blocks within the OPT-1.3B model. We present the Hessian distribution results for both the attention and feedforward blocks, with the red portion indicating the top 10\% of the most significant weight distribution. We observed that the salient weights of Q, K, and V in the OPT family tend to concentrate in some columns or rows. Moreover, we noticed that salient weights in the \textit{Out Projection} layer of multi-head self-attention blocks are distinctly concentrated in specific columns, supporting our structured selection approach discussed in the main text. In contrast, the distribution of salient weights in the feedforward layers is more dispersed. Based on these observations, we adopt a sensitivity-based structured search method to identify salient columns.


\end{document}